%% file: main.tex
\documentclass[journal]{IEEEtran}
\usepackage{amsmath}

\DeclareMathOperator*{\argmin}{arg\,min}
\usepackage{amssymb}
\usepackage{color}
\usepackage{mathrsfs}
\usepackage{mathtools}
\DeclarePairedDelimiter{\norm}{\lVert}{\rVert}
\usepackage[ruled,linesnumbered]{algorithm2e}
\usepackage{makecell}
\input{definitions}

\usepackage{cite}

\usepackage[pdftex]{graphicx}

\ifCLASSINFOpdf
\else
\fi
\usepackage{algorithmic}

\begin{document}
%
\title{Model Compression using Progressive Channel Pruning}
%
%
%

\author{Jinyang~Guo,
        Weichen~Zhang,
        Wanli~Ouyang,~\IEEEmembership{Senior Member,~IEEE,}
        Dong~Xu,~\IEEEmembership{Fellow,~IEEE,}

\thanks{Jinyang Guo, Weichen Zhang, Wanli Ouyang and Dong Xu are with the School of Electrical and Information Engineering, University of Sydney, Sydney, NSW, 2008 Australia. E-mail: \{jinyang.guo, weichen.zhang, wanli.ouyang, dong.xu\}@sydney.edu.au. Corresponding author: Dong Xu.

This work was supported by the Australian Research Council (ARC) Future Fellowship under Grant FT180100116. Wanli Ouyang was supported by the ARC Discovery Project under Grant DP200103223.
}}


%
%

\markboth{IEEE TRANSACTIONS ON CIRCUITS AND SYSTEMS FOR VIDEO TECHNOLOGY, ~Vol.XX, No.XX, XXXX,XXXX}%
{Shell \MakeLowercase{\textit{et al.}}: Bare Demo of IEEEtran.cls for Journals}
%



\maketitle
\IEEEpubid{\begin{minipage}{\textwidth}\vspace{10mm}\ \\[12pt] \centering
  Copyright © 20xx IEEE. Personal use of this material is permitted. However, permission to use this material for any other purposes must be obtained from the IEEE by sending an email to pubs-permissions@ieee.org.
\end{minipage}}

\begin{abstract}
In this work, we propose a simple but effective channel pruning framework called Progressive Channel Pruning (PCP) to accelerate Convolutional Neural Networks (CNNs). In contrast to the existing channel pruning methods that prune channels only once per layer in a layer-by-layer fashion, our new progressive framework iteratively prunes a small number of channels from several selected layers, which consists of a three-step attempting-selecting-pruning pipeline in each iteration. In the attempting step, we attempt to prune a pre-defined number of channels from one layer by using any existing channel pruning methods and estimate the accuracy drop for this layer based on the labelled samples in the validation set. In the selecting step, based on the estimated accuracy drops for all layers, we propose a greedy strategy to automatically select a set of layers that will lead to less overall accuracy drop after pruning these layers. In the pruning step, we prune a small number of channels from these selected layers. We further extend our PCP framework to prune channels for the deep transfer learning methods like Domain Adversarial Neural Network (DANN), in which we effectively reduce the data distribution mismatch in the channel pruning process by using both labelled samples from the source domain and pseudo-labelled samples from the target domain. Our comprehensive experiments on two benchmark datasets demonstrate that our PCP framework outperforms the existing channel pruning approaches under both supervised learning and transfer learning settings. 
\end{abstract}


%
\IEEEpeerreviewmaketitle

\section{Introduction}
\label{Sec:Intro}
%
%
%
%

\IEEEPARstart{W}{hile} deep learning technologies have been successfully used for many computer vision tasks, it is still a challenging task to deploy deep neural networks on mobile devices due to tight computation resources and limited battery power. Several model compression approaches (see Section \ref{Sec:RelatedWork} for more details) have been recently developed to deploy deep models on resource-constrained devices, among which channel pruning technologies are attracting increasing attention as these technologies are often efficient on both CPUs and GPUs without requiring special implementation. 

In this work, we propose a new iterative channel pruning framework called Progressive Channel Pruning (PCP) for model compression under both supervised and transfer learning settings. Given any trained Convolutional Neural Networks (CNNs), in Section \ref{Sec:PCPSupervised}, we introduce our PCP framework under supervised learning setting, which iteratively prunes a small number of channels from several selected layers. Specifically, our framework consists of a three-step attempting-selecting-pruning pipeline in each iteration. In the first attempting step, we attempt to prune a pre-defined number of channels from one layer by using any existing channel pruning methods (e.g., \cite{he2017channel, luoiccv2017}). As a result, we can estimate the accuracy drop after pruning this layer based on the labelled samples in the validation set. In the second selecting step, we propose a greedy strategy to automatically select a set of layers by using the estimated accuracy drops for all layers obtained in the attempting step, where the overall accuracy drop is the smallest after pruning these selected layers. In the third pruning step, we prune a small number of channels from these selected layers by using the existing channel pruning approaches. The above three steps are repeated until the number of FLOPs or the compression ratio is satisfied. 

In Section \ref{Sec:PCPtransfer}, we further extend our PCP framework to prune channels for the deep transfer learning methods like Domain Adversarial Neural Network (DANN), in which we use both labelled samples from the source domain and pseudo-labelled samples from the target domain in the channel pruning process. To obtain pseudo-labelled target samples, we firstly use the DANN approach to fine-tune the pre-trained model and obtain the initial deep model to be pruned. We then employ this initial model to predict pseudo-labels for target samples. By additionally using pseudo-labelled target samples, our PCP framework can effectively reduce the data distribution mismatch between two domains in the channel pruning process. 

In Section \ref{Sec:Experiment}, we perform comprehensive experiments on two benchmark datasets: ImageNet~\cite{imagenet} and Office-31~\cite{saenko2010adapting} to compare our PCP framework with the existing channel pruning methods under both supervised learning and transfer learning settings. The results clearly demonstrate the effectiveness of our PCP approach for model compression under both settings.  

We would like to highlight the main contributions of this work. First, to the best of our knowledge, this is the first channel pruning framework that can automatically decide the network structure (i.e., the optimal number of remained channels at each layer) after channel pruning for both supervised learning models (e.g., VGG, AlexNet and ResNet) and unsupervised domain adaptation (UDA) methods like DANN~\cite{ganin2016domain}. Several existing channel pruning technologies (e.g., \cite{he2017channel,luoiccv2017}) can be readily incorporated in both the attempting and the pruning steps of our PCP framework. Second, we can simultaneously obtain a series of compressed deep models at different compression ratios, which well suits the scalable applications where the compression ratios may change over time. For example, if we aim to prune a model to achieve 5$\times$ compression ratio, we can obtain a series of compressed deep models at lower compression ratios such as 4$\times$ and 2$\times$ as the by-products in our PCP framework without running extra experiments with different settings. Finally, our PCP method achieves promising model compression results under both supervised learning and unsupervised domain adaptation settings.

\section{Related Work}
\label{Sec:RelatedWork}

Our work is related to model compassion and transfer learning. We summarize the most related works below. 

\subsection{Model Compression and Acceleration}
\label{Sec:ModelCompression}
A large number of model compression methods are proposed to accelerate neural networks, which can be roughly grouped into five categories: tensor factorization \cite{lebedev2014factorization,jaderberg2014factorization,kim2015factorization,gong2014factorization,xue2013factorization}, quantization \cite{rastegari2016quantization,guo2024compressing,lv2024ptq4sam,yang2024llmcbench}, optimized implementation \cite{bagherinezhad2017lcnn,lavin2016fast,mathieu2014fast,vasilache2014fast}, compact network design \cite{howard2017mobilenet,zhang2017shufflenet,liu2024lta-pcs}, and connection pruning \cite{han2016eie,he2017channel,luoiccv2017,hu2016nettrim,li2017filter,molchanov2017taylor,han2015learning,guo2020multi,guo2023multidimensional,guo2020channel,wang2024ptsbench,guo2021jointpruning,guo2023cbanet,guo20223d}. 

In \textbf{tensor factorization}~\cite{lebedev2014factorization,jaderberg2014factorization,kim2015factorization,gong2014factorization,xue2013factorization}, the weights are decomposed into light-weight pieces. For example, a 3$\times$3 convolutional filter is decomposed as one 1$\times$3 and one 3$\times$1 filters in \cite{jaderberg2014factorization}. In \cite{xue2013factorization, dentonnips2014, girshickcvpr2015fastrcnn}, truncated SVD is used to accelerate the fully connected layers. \textbf{Quantization} technology in \cite{rastegari2016quantization} reduces the computational complexity by directly representing floating point values with lower bits. The \textbf{compact network design} works~\cite{howard2017mobilenet,zhang2017shufflenet} change the conventional convolution into different formats. For example, group-wise convolution is used in \cite{zhang2017shufflenet} to accelerate the conventional convolution. \textbf{Optimized implementation} technologies~\cite{bagherinezhad2017lcnn,lavin2016fast,mathieu2014fast,vasilache2014fast} are also proposed to use special convolution algorithms (e.g., FFT) to accelerate the convolution operation.

For \textbf{connection pruning}, Han et al.~\cite{han2015learning} proposes a new iterative approach to construct a sparse network by removing all connections where the corresponding weights are lower than a pre-defined threshold. However, this work often achieves poor practical acceleration performance due to cache and memory access issues. To address this issue, many works (e.g., \cite{hu2016nettrim,li2017filter,molchanov2017taylor,zhao2019variational,lin2019towards,peng2019collaborative}) are proposed to reduce redundant connections at filter level. Luo et al.~\cite{luoiccv2017} proposes a channel pruning approach, in which less important channels at each layer are removed based on the statistics information computed from the next layer. He et al.~\cite{he2017channel} proposes a two-step approach to prune each layer by optimizing a LASSO regression problem for channel selection and a least square problem for updating the corresponding weights. More recently, He et al.~\cite{heeccv2018} proposes to leverage reinforcement learning to improve model compression results by automatically searching the model design space. Zhao et al.~\cite{zhao2019variational} uses variational technique to prune filters in networks and adversarial learning technology is used in \cite{lin2019towards} to prune redundant structures. Inter-channel relationship is used in \cite{peng2019collaborative} to determine which channel should be pruned/reserved. The work \cite{chen2019cooperative} proposes a cross-domain pruning strategy to simultaneously prune both the source and the target networks. Compared with \cite{he2017channel,luoiccv2017,heeccv2018,chen2019cooperative}, our work can automatically determine network structure (i.e., the optimal number of remained channels at each layer). More importantly, the existing methods \cite{heeccv2018, he2017channel, luoiccv2017, zhao2019variational, lin2019towards, peng2019collaborative} do not study the more challenging channel pruning problem under the UDA setting, where the manually designed/tuned network structure or automatically learned network structure (based on the source domain data only) is not the optimal one due to the considerable data distribution mismatch between source and target domains. As a result, the existing works cannot achieve satisfactory performance under the UDA setting. In contrast to \cite{heeccv2018,he2017channel,luoiccv2017,zhao2019variational,lin2019towards,peng2019collaborative}, our PCP framework can gradually learn the optimal network structure and reduce the data distribution mismatch by utilizing both labelled source samples and pseudo-labelled target samples in an iterative fashion and its effectiveness is also demonstrated by our comprehensive experiments.

\subsection{Deep Transfer Learning}
\label{Sec:DeepTransferLearning}
The deep transfer learning methods aim to learn domain invariant features by explicitly reducing the data distribution mismatch between the source domain and the target domain, which can be roughly categorized as statistic-based approaches~\cite{long2015learning,long2016unsupervised,sun2016deep,tzeng2014deep} and adversarial learning based approaches~\cite{bousmalis2017unsupervised,bousmalis2016domain,ganin2015unsupervised,ganin2016domain,liu2016coupled,tzeng2017adversarial, zhang2018collaborative}. Domain Adversarial Neural Network (DANN)~\cite{ganin2016domain} is one of the most representative works, in which domain-invariant features are learned by inversely back-propagating the gradients from the loss of the domain classifier. In this work, we take DANN as an example to demonstrate the effectiveness of our PCP framework for model compression. To the best of our knowledge, there is no existing method that can automatically decide the network structure to accelerate deep transfer learning methods like DANN in order to deploy them on resource-constrained devices.

\section{Progressive Channel Pruning under supervised learning setting}
\label{Sec:PCPSupervised}

Several channel pruning method \cite{luoiccv2017,he2017channel} can be readily used in our framework. We choose the approach in \cite{he2017channel} to demonstrate the effectiveness of our PCP framework because its source code is publicly available and its performance is promising. In this section, we firstly review this channel pruning method \cite{he2017channel}. Then we introduce our proposed Progressive Channel Pruning (PCP) method under the supervised learning setting.

\subsection{Brief Introduction of Channel Pruning}
\label{Sec:ChannelPruning}
In order to reduce the parameters and computation, the channel pruning methods reduce the number of input channels at each layer in a CNN model. More formally, we consider pruning one convolutional layer as an example. Assume the convolutional filters $\mathbf{W} \in \mathbb{R}^{n \times c \times h \times w}$ are applied to the input feature volume $\mathbf{X} \in \mathbb{R}^{N \times c \times h \times w}$ to produce the output matrix $\mathbf{Y} \in \mathbb{R}^{N \times n}$, where $N$ denotes the number of samples. $c$ and $n$ denote the number of input and output channels of this layer, respectively. $h$ and $w$ are the height and width of the filters, respectively. For channel pruning, we aim to reduce the number of channels for the feature volume $\mathbf{X}$ from $c$ to $c'$ ($0 \leq c' \leq c$) and remove the corresponding channels of the filters at this layer and the corresponding filters in the previous layer.

The channel pruning method~\cite{he2017channel} formulates the channel pruning problem as follows:
\begin{eqnarray}
\begin{aligned}
\label{eqn:ls0}
\argmin_{\mathbf{W}, \bm{\beta}} \cL_R(\mathbf{Y}, \mathbf{X}, \mathbf{W}, \bm{\beta}) \textrm{, subject to } 
||\bm{\beta}||_0 \leq c', \\
\textrm{where } \cL_R(\mathbf{Y}, \mathbf{X}, \mathbf{W}, \bm{\beta})= \norm*{ \mathbf{Y}-\sum^{c}_{i=1}\beta_{i}\mathbf{X}_i\mathbf{W}_i^\top}_{F}^{2}.
\label{eqn:ls01}
\end{aligned}
\end{eqnarray}
$\mathbf{X}_{i} \in \mathbb{R}^{N\times hw}$ is the reshaped matrix from the shape of $N \times h \times w$, which is sliced from the $i$-th channel of the input feature volume $\mathbf{X}$, $i=1, \ldots, c$. $\mathbf{W}_{i} \in \mathbb{R}^{n\times hw}$ is the reshaped matrix from the shape of $n \times h \times w$, which is sliced from the $i$-th channel of the convolutional filters. $\bm{\beta} = \{\beta_{1}, \dots, \beta_{c}\} \in \mathbb{R}^{c}$ is a channel selection vector, namely, if its $i$-th entry $\beta_{i}=0$, the corresponding channels in $\mathbf{X}$ and $\mathbf{W}$ will be removed from this layer. $\norm*{\cdot}_{F}$ is the Frobenius norm and $\norm*{\cdot}_0$ is the $l_0$ norm. For better presentation, the bias term and the activation function in the convolutional layer are not included in Eq. (\ref{eqn:ls0}). 

In order to solve the optimization problem in Eq.~(\ref{eqn:ls0}), two sub-problems are alternatively optimized. First, we fix the parameters of the convolutional filters $\mathbf{W}$, and solve $\bm{\beta}$ for channel selection by formulating it as a LASSO regression problem as follows:
\begin{eqnarray}
\begin{aligned}
\label{eqn:lasso}
\bar{\bm{\beta}} &= \argmin_{\bm{\beta}} \cL_R(\mathbf{Y}, \mathbf{X}, \mathbf{W}, \bm{\beta}) + \lambda \| \bm{\beta} \|_{1} \\
&= \argmin_{\bm{\beta}}\ \norm*{ \mathbf{Y}-\sum^{c}_{i=1}\beta_{i}\mathbf{X}_i\mathbf{W}_i^\top}_{F}^{2} + \lambda \|\bm{\beta}\|_1 \label{eqn:lasso1}, \\
&\!\!\!\!\!\text{subject to}\ \|\bm{\beta}\|_{0} \leq c', \\
\end{aligned}
\end{eqnarray}
where $\norm*{\cdot}_1$ is the $l_1$ norm.

Second, we fix $\bm{\beta}=\bar{\bm{\beta}}$ and find the optimal $\mathbf{W}$ by solving a least squares optimization problem as follows:
\begin{eqnarray}
\begin{aligned}
\bar{\mathbf{W}}&=\argmin_{\mathbf{W}} \cL_R(\mathbf{Y}, \mathbf{X}, \mathbf{W}, \bar{\bm{\beta}}) \label{eqn:ls} \\
&= \argmin_{\mathbf{W}}\ \norm*{ \mathbf{Y}-\sum^{c}_{i=1}\bar{\beta}_{i}\mathbf{X}_i\mathbf{W}_i^\top}_{F}^{2},
\end{aligned}
\end{eqnarray}
where $\bar{\bm{\beta}} = \{\bar{\beta_{1}}, \dots, \bar{\beta_{c}}\} \in \mathbb{R}^{c}$ is the channel selection vector obtained by solving Eq.~(\ref{eqn:lasso}) and $\bar{\beta}_i$ is the $i$-th entry of $\bar{\bm{\beta}}$.

We gradually increase $\lambda$ from 0 at the first iteration and alternatively solve Eq.~(\ref{eqn:lasso}) and Eq.~(\ref{eqn:ls}) to update the channel selection vector and the corresponding weights until the constraint $ \|\bm{\beta}\|_{0} \leq c'$ is satisfied. 

\subsection{Pruning Residual Architecture}
\label{Sec:CPResNet}

\begin{figure}[t]
\begin{center}
\includegraphics[width=\linewidth]{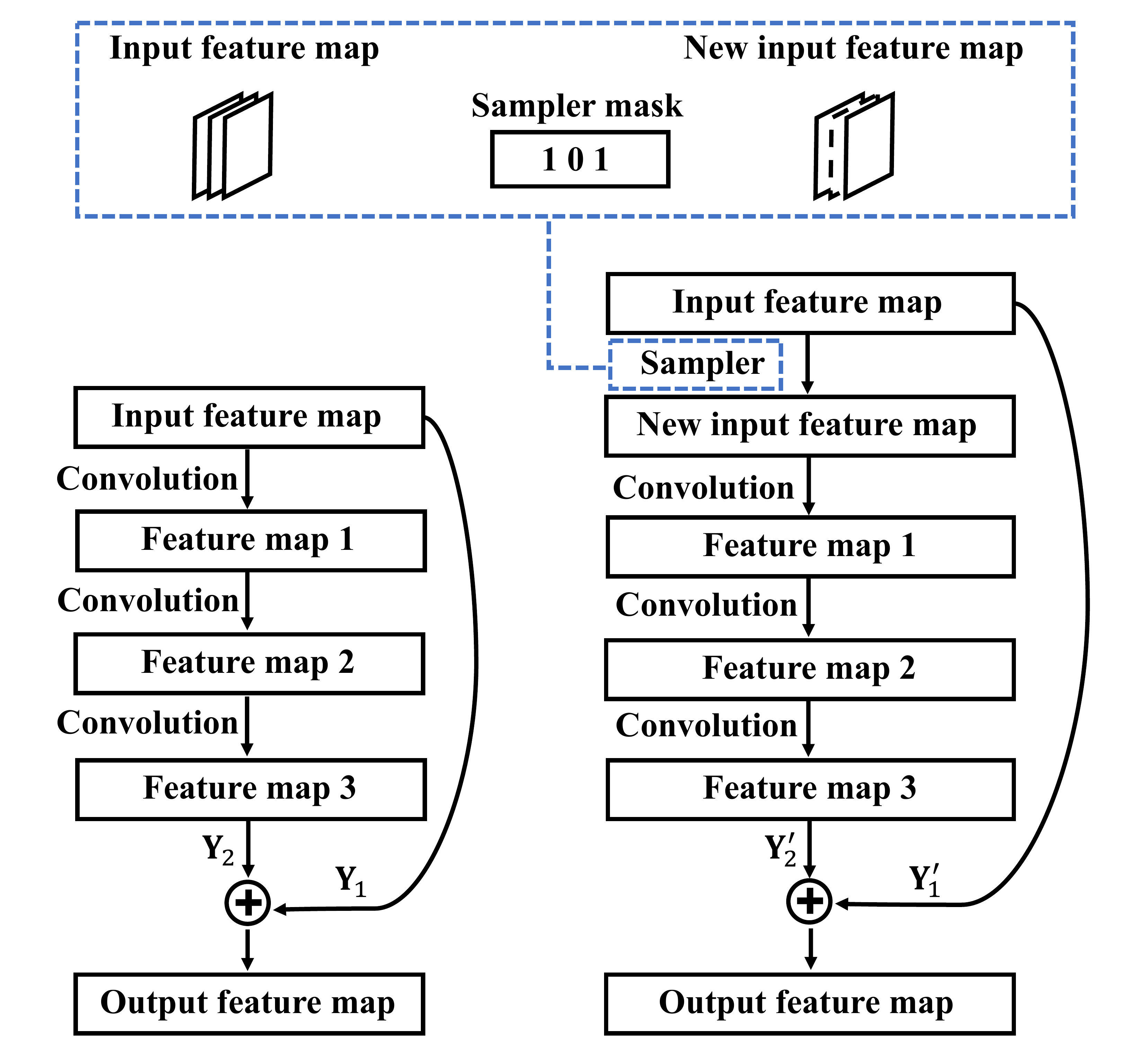}
\end{center}
\vspace{-14pt}
\caption{Illustration of pruning one residual block. Left: Original residual block. Right: Pruned residual block with all previous blocks pruned. $\mathbf{Y}_{1}$ and $\mathbf{Y}_{2}$ are the input feature map of the first convolutional layer and the output feature map of the third convolutional layer in the residual block in the original model, respectively. $\mathbf{Y}_{1}^{'}$ and $\mathbf{Y}_{2}^{'}$ are the input feature map of the first convolutional layer and the output feature map of the third convolutional layer after pruning the previous layers, respectively. For pruning the first convolutional layer, we add a sampler to avoid directly pruning the input feature map to ensure the shapes of two inputs of the summation operation are the same. The dotted box of this figure illustrates a sampler for an input feature map with three channels. When the sampler mask is 0, the corresponding channel in the input feature map will be removed. For pruning the third convolutional layer, we change the optimization goal in Eq.~(\ref{eqn:ls01}) from $\mathbf{Y}$ (Note $\mathbf{Y}=\mathbf{Y}_{2}$ when pruning the third convolutional layer) to $\mathbf{Y}_{1} + \mathbf{Y}_{2}-\mathbf{Y}_1^{'}$ to avoid error accumulation.}
\label{fig:resnet}
\end{figure}

Due to the shortcut design in the residual block, we cannot directly prune the first convolutional layer in the residual block as described above because the input feature of the first convolutional layer will be added to the output feature of the third convolutional layer in this residual block. Channel pruning method~\cite{he2017channel} develops several strategies to prune the residual network. We choose two of them to prune our residual network, which are illustrated in Fig.~\ref{fig:resnet}.

\textbf{Pruning the first convolutional layer.} As shown in Fig.~\ref{fig:resnet}, we add a channel sampler to obtain a feature map with reduced channels as the new input feature map of the subsequent convolutional layers in this residual block to avoid directly pruning the input feature map in Fig.~\ref{fig:resnet} to ensure there is no shape mismatch for the summation operation later. We follow \cite{he2017channel} to solve the LASSO optimization problem in Eq.~(\ref{eqn:lasso}) to generate the sampler mask. 

\textbf{Pruning the third convolutional layer.} As shown in Fig.~\ref{fig:resnet} (right), the output feature map is the summation of $\mathbf{Y}_{1}^{'}$ and $\mathbf{Y}_{2}^{'}$, where $\mathbf{Y}_{1}^{'}$ and $\mathbf{Y}_{2}^{'}$ are the input feature map of the first convolutional layer and the output feature map of the third convolutional layer after pruning the previous layers, respectively. Unlike pruning the residual branch (the branch with Feature map 1, Feature map 2 and Feature map 3 in Fig.~\ref{fig:resnet}), we cannot approximate $\mathbf{Y}_1$ by solving the optimization problem in Eq.~(\ref{eqn:ls01}) because the shortcut branch in the residual block is parameter-free. Therefore, the error generated by the shortcut will be accumulated when the network is deep. To compensate this error, we change our optimization goal from $\mathbf{Y}$ (Note $\mathbf{Y}=\mathbf{Y}_2$ when pruning the third convolutional layer) in Eq.~(\ref{eqn:ls01}) to $\mathbf{Y}_1 + \mathbf{Y}_{2}-\mathbf{Y}_1^{'}$, where $\mathbf{Y}_1$ and $\mathbf{Y}_{2}$ are the input feature map of the first convolutional layer and the output feature map of the third convolutional layer in the original model, respectively. $\mathbf{Y}_1^{'}$ is the input feature map of the first convolutional layer after pruning the previous blocks.

We prune the second convolutional layer in the residual branch by using the same approach introduced in Sec. \ref{Sec:ChannelPruning}.

\subsection{Our PCP Approach}
\label{Sec:OurPCP}
\textbf{Notation changes.} In our approach, we consider pruning different layers at different iterations. The superscript $\cdot^{(l)}$ is added to denote the $l$-th layer when necessary (e.g., Eq. (\ref{eqn:obj}) and Algorithm \ref{alg:greedy search}). The subscript $\cdot_{t}$ is added to denote the $t$-th iteration. If not specified, the symbols have the same meaning as those in Eq.~(\ref{eqn:ls0}), Eq.~(\ref{eqn:lasso}) and Eq.~(\ref{eqn:ls}). For example, $\bm{\beta}$ denotes the channel selection vector in Eq.~(\ref{eqn:ls0}) and we will use $\bm{\beta}^{(l)}_t$ to denote the channel selection vector for the $l$-th layer at the $t$-th iteration. We use the subscript $\cdot_0$ to denote the parameters/features in the original model before model compression. For example, $\mathbf{Y}_{0}^{(l)}$ represents the output feature of the $l$-th layer in the original model.

\textbf{PCP approach before the fine-tuning process.}
Given the target compression ratio $R_{target}$, we aim to search the best compressed model $\mathrm{M}_{out}$ with the highest testing accuracy. $R_{target}$ can be the target compression ratio of FLOPs or the number of parameters.
Formally, we define an image classification network as a function $\cN(\cdot)$. Our objective function can be written as:
\begin{equation}
\begin{aligned}
&\argmin_{\tilde{\mathbf{\Theta}}, \mathcal{B}}\cL_c(\cN(\mathcal{I}; \tilde{\mathbf{\Theta}}, \mathcal{B}), y),  \\
&\text{subject to } R(\mathcal{B}) \geq R_{target},
\end{aligned}
\label{eqn:obj}
\end{equation}
where $\cL_c$ denotes the cross-entropy loss. $\mathcal{I}$ is the input image and $y$ is the label for the input image $\mathcal{I}$. $\tilde{\mathbf{\Theta}}$ represents all of the network parameters. $\mathcal{B}=\{\bm{\beta}^{(1)}, \bm{\beta}^{(2)},...,\bm{\beta}^{(N_l)}\}$ is a set of channel selection vectors, where $\bm{\beta}^{(l)} \in \mathbb{R}^{c_{l}}$ is the channel selection vector for the $l$-th layer and $c_{l}$ is the number of channels of the input feature volume for the $l$-th layer. $N_{l}$ is the number of layers of the network. $\cN(\mathcal{I}; \tilde{\mathbf{\Theta}},\mathcal{B})$ is the predicted probability of the input image $\mathcal{I}$ based on the parameters after channel pruning and the channel selection vectors. $R(\mathcal{B})$ is a function of $\mathcal{B}$, which represents the compression ratio. In Eq. (\ref{eqn:obj}), we drop the summation over the training samples for better presentation.

The goal of the task is to improve the classification accuracy of the compressed model by minimizing the cross-entropy loss in Eq. (\ref{eqn:obj}). However, the number of possible solutions for the discrete selection choices of $\mathcal{B}$ makes the exhaustive search intractable. Back-propagation is also required to optimize the parameters $\tilde{\mathbf{\Theta}}$, which is time-consuming. As a result, it is difficult to directly solve the objective function in Eq.~(\ref{eqn:obj}). To address this issue, the channel pruning method \cite{he2017channel} proposes to optimize Eq. (\ref{eqn:ls0}) at each layer in a layer-by-layer fashion. But it is not related to the cross-entropy loss.

\begin{figure}[t]
\begin{center}
\includegraphics[width=\linewidth]{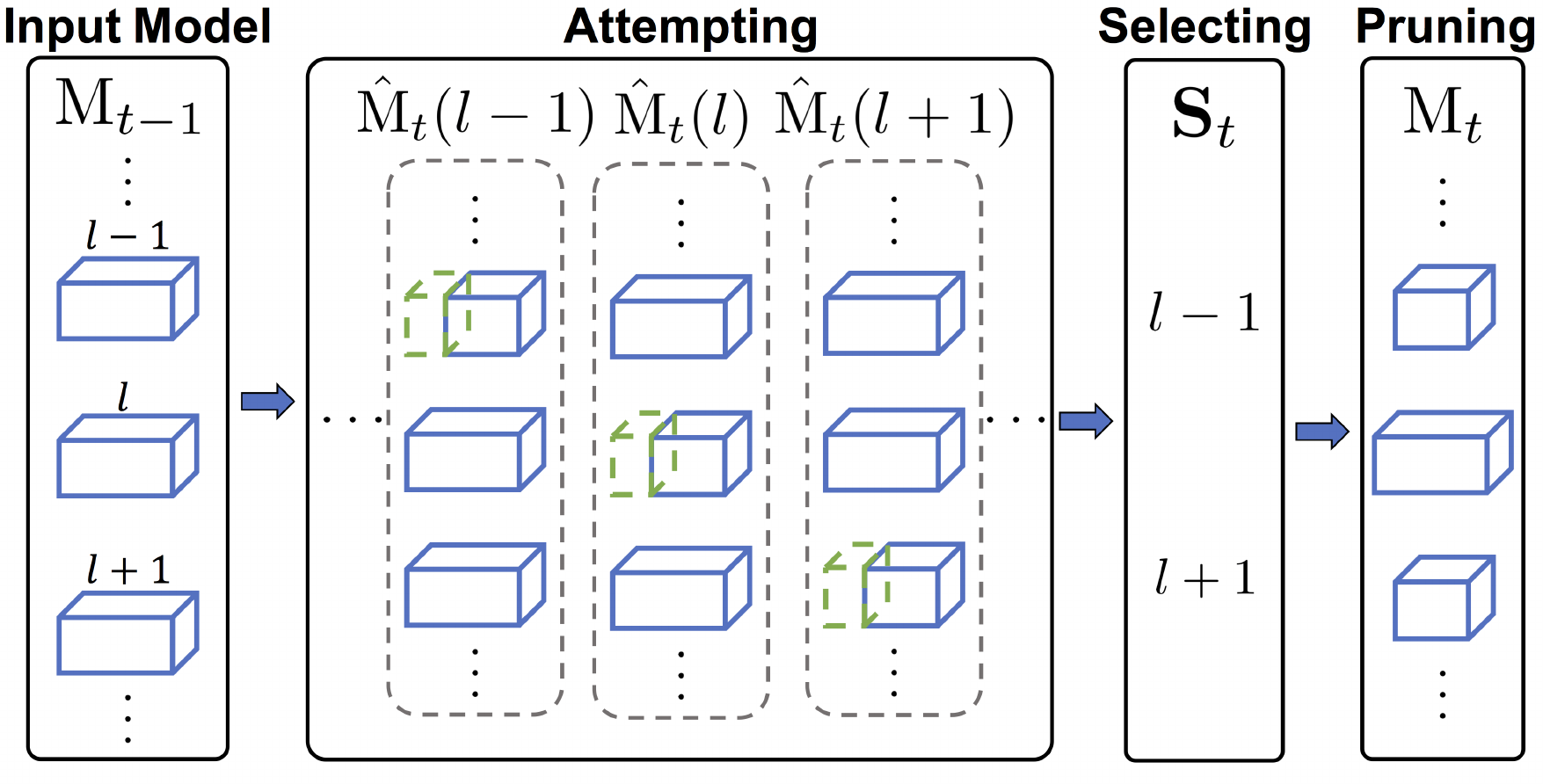}
\end{center}
\vspace{-14pt}
\caption{Illustration of the three-step pipeline in our PCP method at the $t$-th iteration. In this figure, we only illustrate the three-step pipeline for layer ($l-1$), layer $l$ and layer ($l+1$). The input model at the $t$-th iteration is the output model from the previous iteration with its parameters denoted as $\mathrm{M}_{t-1}$. In order to prune a pre-defined number of channels in each layer and evaluate these models on the validation set, we perform the attempting process three times for layer ($l-1$), layer $l$ and layer ($l+1$), respectively. In the selecting step, we set the number of selected layers as 2 in this figure, and the two selected layers are ($l-1$) and ($l+1$) as we have a smaller overall accuracy drop after pruning these two layers. In the pruning step, we prune these two selected layers to obtain the output model at the $t$-th iteration with its parameters denoted as $\mathrm{M}_t$.}
\label{fig:iteration}
\end{figure}

To this end, we propose a greedy search algorithm to solve Eq. (\ref{eqn:ls0}) while taking Eq. (\ref{eqn:obj}) into consideration during the optimization process to obtain a solution that can better approximate the solution of Eq.~(\ref{eqn:obj}). Instead of pruning the channels at each layer only once, we propose to iteratively prune channels, which consists of several steps in each iteration. Specifically, in each iteration, we propose a three-step pipeline including the attempting step, the selecting step and the pruning step.
(1) The attempting step: Given the output model from the previous iteration, we attempt to prune a small number of channels from one layer and estimate the accuracy drop for this layer.
(2) The selecting step: We select a set of optimal layers with minimum accuracy drop after pruning these layers based on the estimated accuracy drops for all layers. 
(3) The pruning step: We prune a small number of channels for the selected layers while keeping the unselected layers unchanged. The weights related to the selected layers are then adjusted accordingly.
The above three steps are repeated for several iterations until the required compression ratio is satisfied. Fig. \ref{fig:iteration} shows the three-step pipeline in one iteration.

\textbf{Notation changes.} Below, we only focus on the operations at the $l$-th layer. For better representation, we ignore the layer index when introducing each operation. However, the layer index will be used when necessary. For example, it is denoted as the superscript $\cdot^{(l)}$ in Algorithm \ref{alg:greedy search}.

(1) The attempting step: At this step, we follow \cite{he2017channel} to attempt to prune a small amount of channels from one layer and estimate the classification accuracy drop of this layer. At the $t$-th iteration,
We optimize a LASSO regression problem for each layer as follows:
\begin{eqnarray}
\label{eqn:lasso_temp}
\begin{aligned}
&\hat{\bm{\beta}}_t = \argmin_{\bm{\beta}} \cL_R(\mathbf{Y}_0, \tilde{\mathbf{X}}_{t-1}, \tilde{\mathbf{W}}_{t-1}, \bm{\beta}) + \lambda \| \bm{\beta} \|_{1}, \\
&\text{subject to}\ \|\bm{\beta}\|_{0} = \|\tilde{\bm{\beta}}_{t-1}\|_{0}-\mathrm{f}_{t}, \\
&\ \ \ \ \ \ \ \ \ \ \ \quad \beta_{i} = 0, \text{if  } \tilde{\beta}_{i, t-1}=0,
\end{aligned}
\end{eqnarray}
where $\mathrm{f}_{t}$ is the pre-defined number of channels to be pruned for the $l$-th layer at the $t$-th iteration. $\beta_{i}$ and $\tilde{\beta}_{i,t-1}$ are the $i$-th element of $\bm{\beta}$ and $\tilde{\bm{\beta}}_{t-1}$, respectively. $\tilde{\bm{\beta}}_{t-1}$ is the channel selection vector obtained from the ($t-1$)-th iteration, which is defined in Eq.~(\ref{eqn:lasso_real}) below.
$\tilde{\mathbf{W}}_{t-1}$ is the adjusted weight obtained from the ($t-1$)-th iteration, which is defined in Eq.~(\ref{eqn:ls_actual}) below. $\tilde{\mathbf{X}}_{t-1}$ is the input tensor at the $l$-th layer in the output model at the ($t-1$)-th iteration with its parameters denoted as $\mathrm{M}_{t-1}$.

Using Eq.~(\ref{eqn:lasso_temp}), we attempt to prune $\mathrm{f}_{t}$ channels from the $l$-th layer and obtain the temporary channel selection vector $\hat{\bm{\beta}}_t$ for the $l$-th layer at the $t$-th iteration. Then we fix $\hat{\bm{\beta}}_t$ and find the temporary weights $\hat{\mathbf{W}}_t$ for the $l$-th layer at the $t$-th iteration by solving the least squares optimization problem as follows:
\begin{eqnarray}
\mathbf{\hat{\mathbf{W}}}_t = 
\argmin_{\mathbf{W}}
\cL_R(\mathbf{Y}_0, \tilde{\mathbf{X}}_{t-1}, \mathbf{W}, \hat{\bm{\beta}}_t). \label{eqn:ls_temp}
\end{eqnarray}
Similar to the channel pruning method~\cite{he2017channel}, we gradually increase $\lambda$ in Eq.~(\ref{eqn:lasso_temp}) and alternatively optimize Eq.~(\ref{eqn:lasso_temp}) and Eq.~(\ref{eqn:ls_temp}) until the constraint is satisfied. The optimization problems in Eq.~(\ref{eqn:lasso_temp}) and Eq.~(\ref{eqn:ls_temp}) are based on the original output feature $\mathbf{Y}_{0}$ and the input feature from the previous iteration $\tilde{\mathbf{X}}_{t-1}$. Therefore, pruning the channels of one layer is independent from pruning the channels of other layers at each iteration. In this way, the optimization processes for all layers in the current iteration can be performed in parallel to improve the efficiency of our PCP framework and will be studied in our future work.

Empirically, we find a model with higher accuracy before the fine-tuning process usually leads to better performance after the fine-tuning process \cite{he2017channel, heeccv2018}. Therefore, we evaluate each pruned model based on our validation set before the fine-tuning process to estimate the accuracy of this model after the fine-tuning process. This accuracy provides an indicator of the overall cross-entropy loss $\cL_c$ in Eq. (\ref{eqn:obj}). Specifically, a higher accuracy represents a smaller cross-entropy loss, which indicates that the pruned channels have a smaller influence on the overall cross-entropy loss, namely, there are more redundancies in this layer.

(2) The selecting step: After the attempting step, we obtain the estimated accuracy drops for all layers, we select $top\_n$ layers with lower accuracy drops. The set of selected layers is denoted as $\mathbf{S}_t$ at the $t$-th iteration. Then, we sort the indices of these layers in ascending order to make sure that we will prune the channels from shallower layers (i.e., close to the input image) to deeper layers (i.e., close to the network output) in the pruning step introduced below.

{\SetAlgoNoLine
\begin{algorithm}[t]
  \caption{Our PCP Algorithm.}  
  \label{alg:greedy search}  
    \KwIn{\\$\mathbf{L}$: The set of layer indices to be evaluated in the attempting step. \\
        The original model: $\mathrm{M}_{0}=\{\bm{\theta}_{0}^{(1)}, \bm{\theta}_{0}^{(2)},\ldots, \mathbf{\Theta}\}$ where $\bm{\theta}_{0}^{(l)}=\{\mathbf{W}_{0}^{(l)}, \bm{\beta}_{0}^{(l)}\}$ for $l \in \mathbf{L}$ and $\mathbf{\Theta}$ is the parameters for other layers that will not be pruned (e.g., fully connected layer). \\
    $top\_n$: The number of selected layers in the selecting step. \\
    $R_{target}$: The target compression ratio.}
    \KwOut{A series of output models with their parameters $\{\mathrm{M}_{1}, \mathrm{M}_{2}, \ldots \}$}
    $t = 1$; \\
    \While {True}
    {
        \For {$l \in \mathbf{L}$ }
        {
            // Attempt to prune the $l$-th layer of the output model from the previous iteration with the parameters $\mathrm{M}_{t-1}$.\\
            $\hat{\bm{\theta}}^{(l)}_t=\{\hat{\mathbf{W}}^{(l)}_t,\hat{\bm{\beta}}^{(l)}_t\}$ by solving Eq. (\ref{eqn:lasso_temp}) and (\ref{eqn:ls_temp}); \\
            // Estimate classification accuracy drop.\\
            Test the model $\hat{\mathrm{M}}_t(l) = \{\bm{\theta}_{t-1}^{(1)}, \ldots, \bm{\theta}_{t-1}^{(l-1)}, \hat{\bm{\theta}}_{t}^{(l)}, \bm{\theta}_{t-1}^{(l+1)}, \ldots , \mathbf{\Theta}\}$ on the validation set and obtain the accuracy $acc_{t}(l)$; \\
        }
        // The selecting step.\\
        Select the $top\_n$ layers with the smallest accuracy drops based on $\{acc_{t}(1), acc_{t}(2), \ldots\}$ and save the indices of the selected layers in ascending order as $\mathbf{S}_{t}$;\\
        // Prune the layers in $\mathbf{S}_t$ from shallower layers to deeper layers.\\
        \For {$p \in \mathbf{S}_t$}
        {
            $\bm{\theta}^{(p)}_t$=$\{\tilde{\mathbf{W}}^{(p)}_t, \tilde{\bm{\beta}}^{(p)}_t\}$ by solving Eq.(\ref{eqn:lasso_real}) and (\ref{eqn:ls_actual});
        }
        \For {$p' \not\in \mathbf{S}_t$ and $p' \in \mathbf{L}$}
        {
            $\bm{\theta}^{(p')}_t = \bm{\theta}^{(p')}_{t-1}$
        }
        $\mathrm{M}_t= \{\bm{\theta}^{(1)}_t, \bm{\theta}^{(2)}_t, \ldots, \mathbf{\Theta}\} $;\\
        // Break when compression ratio reaches $R_{target}$. \\
        $R_{t} = FLOPs(\mathrm{M}_{0})/FLOPs(\mathrm{M}_{t})$; \\
        \If {$R_{t} \geq R_{target}$ }
        {
            Break; \\
        }
        $t = t + 1$; \\
    }
\end{algorithm} 
}

(3) The pruning step: After the set of selected layers $\mathbf{S}_t$ is obtained, we follow \cite{he2017channel} to prune these selected layers in a shallow-to-deep order by solving the optimization problem as follows for $l \in \mathbf{S}_t$: 
\begin{eqnarray}
\label{eqn:lasso_real}
\begin{aligned}
&\tilde{\bm{\beta}}_t = \argmin_{\bm{\beta}} \cL_R(\mathbf{Y}_0, \mathbf{X}_{t}, \tilde{\mathbf{W}}_{t-1}, \bm{\beta})+ \lambda \| \bm{\beta} \|_{1}, \\
&\text{subject to}\ \|\bm{\beta}\|_{0} = \|\tilde{\bm{\beta}}_{t-1}\|_{0}-\mathrm{f}_{t}, \\
&\ \ \ \ \ \ \ \ \ \ \ \quad \beta_{i,t} = 0, \text{if  } \tilde{\beta}_{i, t-1}=0.
\end{aligned}
\end{eqnarray}
$\mathbf{X}_{t}$ is the input tensor of the $l$-th layer at the $t$-th iteration when the previous layers are pruned. In this way, we update the channel selection vector for the selected layers in $\mathbf{S}_{t}$ but keep the channel selection vector unchanged for the unselected layers.

To adjust the network parameters, we solve the least squares optimization problem for the selected layers in $\mathbf{S}_t$ as follows:
\begin{eqnarray}
\tilde{\mathbf{W}}_t=
\argmin_{\mathbf{W}} \cL_R(\mathbf{Y}_0, \mathbf{X}_{t}, \mathbf{W}, \tilde{\bm{\beta}}_t).
\label{eqn:ls_actual}
\end{eqnarray}
In this way, we update the parameters for the selected layers with fixed channel selection vector $\tilde{\bm{\beta}}_t$. We keep the parameters of unselected layers unchanged to save the computational time. 

As an example, suppose the set of selected layers $\mathbf{S}_t=\{2, 4, 6\}$ at the $t$-th iteration. In the pruning step, we use the LASSO solver to solve Eq.~(\ref{eqn:lasso_real}) to update the channel selection vector and then solve Eq.~(\ref{eqn:ls_actual}) to update the parameters of this layer. This process is performed for each layer from shallower layers to deeper layers because the LASSO and least squares optimization problems in Eq.~(\ref{eqn:lasso_real}) and Eq.~(\ref{eqn:ls_actual}) are based on the current feature maps $\mathbf{X}_{t}$. Pruning the shallower layer (e.g., layer 2) will affect the input feature after this layer (e.g., layer 6). This shallow-to-deep sequential updating process can make the deeper layers aware of the change of their input feature caused by pruning the shallower layers, which will lead to more accurate channel selection and better parameter adjustment in deeper layers and it is also used in \cite{he2017channel}.

We prune the channels for several iterations until the model reaches the target compression ratio. Naturally, we can save the models after each iteration to obtain the models with different compression ratios to suit the scalable applications, where the compression ratio may change over time. Algorithm \ref{alg:greedy search} shows the pseudo-code of our PCP algorithm before the fine-tuning process. 

\textbf{Fine-tuning.}
We fine-tune the compressed model based on all labelled training data. Note that the fine-tuning process is not used during the attempting-selecting-pruning process to achieve fast speed.

\section{Progressive Channel Pruning under transfer learning setting}
\label{Sec:PCPtransfer}
In this section, we extend our PCP method to compress deep transfer learning methods, in which we follow the most common unsupervised domain adaptation setting in this task. We take the DANN method~\cite{ganin2016domain} as an example to introduce our PCP approach, which can be readily extended to compress other deep transfer learning methods.

In unsupervised domain adaptation, we have a set of labelled samples from the source domain and a set of unlabelled samples from the target domain, in which the data distributions between the source and target domains are different. The goal of the task is to learn a classifier for the target domain, which can categorize the unlabelled target samples into the correct classes. 

In this task, the progressive channel pruning process consists of three main steps: (1) We firstly pre-train an initial deep transfer learning model (e.g., the DANN model~\cite{ganin2016domain}) by using the samples from both the source domain and the target domain. (2) We compress the initial model by using our PCP framework to accelerate the network while maintaining domain adaptation ability. (3) We fine-tune the compressed model by using the samples from both domains.

\textbf{Pre-train the initial model.}
In order to make the compressed model has a better performance on the target domain, we find it is useful to prune the channels from a deep transfer learning model (e.g., a DANN model) instead of pruning the channels from the deep models before domain adaptation (e.g., the model pre-trained on the ImageNet dataset). Specifically, we use the simple and effective domain adaptation method DANN~\cite{ganin2015unsupervised, ganin2016domain} to obtain the initial model to be compressed. The DANN method uses the same CNN structure like VGG~\cite{simonyan2014very} or ResNet~\cite{he2016deep} to extract features. Nevertheless, it also contains a domain classifier to decide whether the samples are from the source or the target domain and employs a gradient reversal layer to inversely back-propagate the gradients from the domain classifier to extract the domain invariant features. 

\textbf{Channel pruning by using PCP.}
After obtaining the initial DANN model, we use our PCP method to progressively prune the channels for the initial DANN model. Considering that there are labelled source samples and unlabelled target samples in this setting, we propose two strategies to effectively utilize these samples. 
(1) We select a set of highly confident target samples and assign pseudo-labels to them. (2) We choose informative features at suitable spatial locations. These two strategies are introduced below.

First, due to the data distribution mismatch between two domains,  we may achieve poor classification performance on the target domain if we simply estimate the accuracy drops in the attempting step by only using the labelled source data. Therefore, it is beneficial to take the unlabelled target data into consideration. To this end, we use the initial DANN model to predict the categories of the images from the target domain as their pseudo-labels. We progressively prune the channels of the initial DANN model by using Algorithm \ref{alg:greedy search}, in which we use both the labelled source samples and pseudo-labelled target samples in our three-step pipeline.

Second, the original channel pruning method~\cite{he2017channel} uses a small amount of features at some randomly selected spatial positions from the feature maps instead of the whole feature maps when solving the optimization problem in Eq.~(\ref{eqn:ls0}). However, under unsupervised domain adaptation setting, the features at many spatial locations have uninformative responses (e.g., the responses at some spatial location are close to zero in all the channels) because of the large domain variance (e.g., pure color backgrounds in the source domain and complex backgrounds in the target domain). In this case, it is not a good choice to randomly select the spatial locations from the feature maps. To solve this problem, we only use the features at some spatial locations whose variances are larger than an empirically defined threshold across all the channels to solve the optimization problem in the attempting and pruning steps in order to avoids selecting the spatial locations with uninformative features from the background region. 

\textbf{Fine-tuning.}
After finishing the channel pruning process, we fine-tune the pruned model by using the DANN method based on both the labelled source images and the pseudo-labelled target images.

\section{Experiment}
\label{Sec:Experiment}
In this section, we compare our PCP approach with several state-of-the-art channel pruning methods under supervised learning setting on the ImageNet dataset~\cite{imagenet}. We also evaluate our PCP method for the unsupervised domain adaptation task on the Office-31 dataset~\cite{saenko2010adapting}. 

\subsection{Experiments under supervised learning setting}
\label{Sec:ExperimentSupervisedLearning}
We evaluate three popular models: VGG-16~\cite{simonyan2014very}, AlexNet~\cite{krizhevsky2012imagenet} and ResNet-50~\cite{he2016deep} on the ImageNet dataset~\cite{imagenet} as examples to demonstrate the effectiveness of our approach for model compression. 

The ILSVRC-2012 dataset~\cite{imagenet} consists of over one million training images and 50,000 testing images from 1000 categories. We split 50,000 images from the training set as our validation set to evaluate the accuracy drops for each layer at the attempting step. At the fine-tuning stage, we resize the images to the size of 256 $\times$ 256 and random crop the resized images to the size of 224 $\times$ 224. We center crop the resized images to the size of 224 $\times$ 224 at the testing stage. The other implementation details for different network structures and datasets will be discussed in the corresponding sections below.

\begin{table}[t]
\caption{Top-5 accuracies (\%) of different channel pruning methods when compressing the VGG-16 model on the ImageNet dataset under the compression ratios of 2$\times$, 4$\times$ and 5$\times$. The Top-5 accuracy of the VGG-16 model before channel pruning is 89.9\%. For better presentation, we directly report the accuracies after the fine-tuning process. For the channel pruning method~\cite{he2017channel} and the AMC method~\cite{heeccv2018}, we directly quote the results from the papers. For the filter pruning method \cite{li2017filter}, we copy the results from \cite{he2017channel} because the results are not reported in the original paper.}
\begin{center}
\begin{tabular}{|c|c|c|c|}
\hline
Compression Ratios & 2$\times$ & 4$\times$ & 5$\times$ \\
\hline
\makecell{Filter Pruning~\cite{li2017filter}} & \makecell{89.1} & \makecell{81.3} & \makecell{75.3} \\
\hline
\makecell{Channel Pruning~\cite{he2017channel}} & \makecell{89.9} & \makecell{88.9} & \makecell{88.2} \\
\hline
AMC~\cite{heeccv2018} & - & - & \makecell{88.5} \\
\hline
\textbf{Ours} & \makecell{\textbf{90.0}} & \makecell{\textbf{89.3}} & \makecell{\textbf{89.0}} \\
\hline
\end{tabular}
\end{center}
\label{tab:vggimagenet}
\end{table}

\textbf{Results with VGG-16.}
\label{Sec:ExperimentVGG}
VGG-16 is a 16-layer network, which consists of 13 convolutional layers and 3 fully connected layers. Considering that the input of the first convolutional layer is the raw image, we only prune the last 12 convolutional layers of the VGG-16 model (i.e., conv1\_2 to conv5\_3). The compression ratio in this section refers to the number of FLOPs of the original model before model compression over that of the compressed model.

In our experiment, we empirically set the number of selected layers in the selecting step (i.e., $top\_n$) as 4 to accelerate the three-step pipeline in our PCP framework. When the compression ratio is smaller than 2, we set the pre-defined number of channels to be pruned $\mathrm{f}^{(l)}_t$ as max\{30\% of the current number of channels, 40\} to aggressively prune more channels in each iteration as there are more redundancies in the model in this case. When the compression ratio increases from 2 to 5, we gradually decrease $\mathrm{f}^{(l)}_t$ to min\{10\% of the current number of channels, 40\}. It takes 12 iterations to reach $5\times$ compression ratio. In the fine-tuning stage, we set the learning rate as $1e^{-3}$ and use SGD optimizer with the batch size of 128 for optimization. Following the same setting in the baseline method channel pruning~\cite{he2017channel}, we fine-tune the compressed models for 10 epochs. We set other parameters the same as the original VGG work~\cite{simonyan2014very}. 

Table \ref{tab:vggimagenet} shows the results of our PCP framework under the compression ratios of 2$\times$, 4$\times$ and 5$\times$. We compare our method with several state-of-the-art channel pruning approaches including: Filter Pruning (FP)~\cite{li2017filter}, Channel Pruning (CP)~\cite{he2017channel}, and AMC~\cite{heeccv2018}. It is worth mentioning that the CP method needs to carefully design the number of remained channels at each layer of the network. The AMC method uses reinforcement learning to decide the number of remained channels at each layer in a non-progressive fashion. Under the compression ratio of 5$\times$, our PCP approach outperforms FP and CP methods by more than 0.8\% and surpasses the AMC method by 0.5\%. 
Since the baseline methods FP, CP and AMC do not report the Top-1 accuracies, we cannot compare the Top-1 accuracies with them. However, we report the Top-1 accuracies of our PCP framework for better comparison with other approaches. The Top-1 accuracies of our method under 2$\times$, 4$\times$ and 5$\times$ compression ratio are 71.0\%, 69.5\%, and 69.2\%, respectively. We would highlight that our PCP approach is even better than the original VGG-16 model in terms of Top-5 accuracies under the compression ratio of 2$\times$. A possible explanation is that the original VGG-16 model may have redundant channels and it is therefore helpful to discard these redundant channels.

In addition, we compare the running time of our PCP approach with the CP method for compressing the VGG-16 model when using one GTX2080Ti GPU. For the CP method, it takes 0.3 hours for the pruning process and 15 hours for the fine-tuning process. In our PCP framework, it takes 1.4 hours for the pruning process (i.e., our three-step pipeline) and 15 hours for the fine-tuning process. The total running time of the whole framework increases by 7.2\% for our PCP method when compared with that for the baseline CP method, which demonstrates that our PCP framework can automatically decide the remained number of channels for each layer and only requires a small amount of extra time.

\begin{figure}[t]
\begin{center}
  \includegraphics[width=\linewidth]{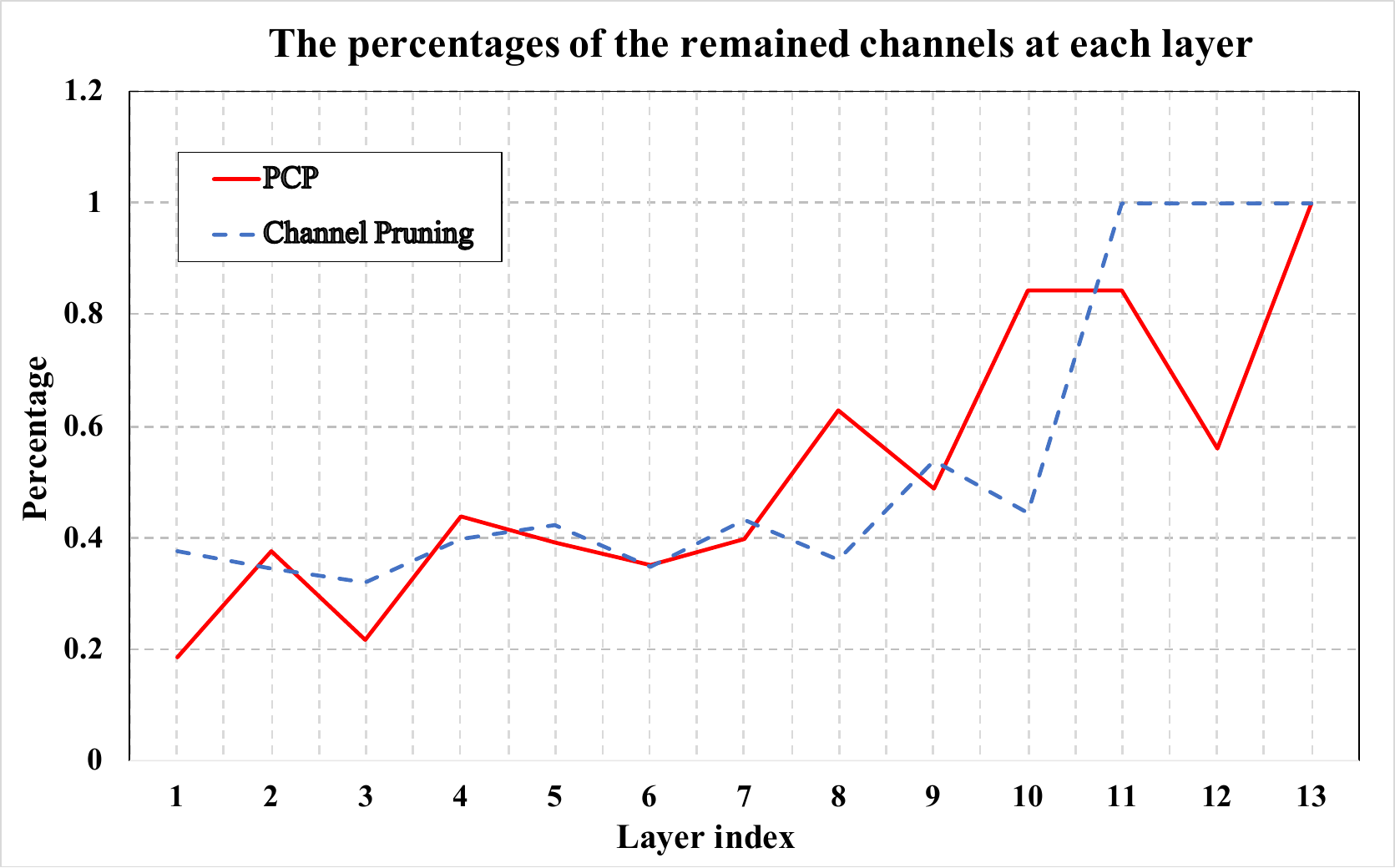}
\end{center}
\vspace{-14pt}
\caption{Percentages of the remained channels (i.e., the number of remained channels after channel pruning over the original number of channels before channel pruning) at each layer in the channel pruning method~\cite{he2017channel} and our PCP framework under 5$\times$ compression ratio of the VGG-16 model on the ImageNet dataset. The layer indices along the horizontal axis indicate the inputs of last 12 convolutional layers of the VGG-16 model (i.e., conv1\_2 to conv5\_3). The percentages of the remained channels automatically decided by our PCP framework has a similar trend with that in the channel pruning method~\cite{he2017channel}.}
\label{fig:density}
\end{figure}

We also report the Top-5 accuracies of our PCP method after the fine-tuning process by using different number of selected layers in the selecting step (i.e., $top\_n$'s). When setting $top\_n$ as 1, 2, 4, 6, the Top-5 accuracies of our PCP method after the fine-tuning process are 88.8\%, 88.9\%, 89.0\%, 88.7\% for the compressed VGG-16 model under 5$\times$ compression ratio, respectively. From the results, the accuracies of our PCP method increase when $top\_n$ increases from 1 to 4 and then decrease when $top\_n$ is higher than 4. 
When setting the hyper-parameter $\mathrm{f}^{(l)}_{t}$, we also perform the experiments by using different $\mathrm{f}^{(l)}_{t}$. When $\mathrm{f}^{(l)}_{t}$ are max\{20\% of the current number of channels, 40\}, max\{30\% of the current number of channels, 40\} and max\{40\% of the current number of channels, 40\}, the Top-5 accuracies after the fine-tuning process are 89.2\%, 89.0\%, and 88.8\% for the compressed VGG-16 model under 5$\times$ compression ratio, respectively. Increasing $\mathrm{f}^{(l)}_{t}$ will increase the speed of our PCP framework but decrease the accuracies of the compressed model. Therefore, we set $\mathrm{f}^{(l)}_{t}$ as max\{30\% of the current number of channels, 40\} in our experiment for the trade-off between the performance and the speed. From the results with different choices of $top\_n$ and $\mathrm{f}^{(l)}_{t}$, we can observe that our PCP framework is not sensitive to these hyper-parameters.

In Fig. \ref{fig:density}, we take the VGG-16 model under 5$\times$ compression ratio as an example to report the percentage of the remained channels at each layer for our method and the CP method~\cite{he2017channel}, which is the number of remained channels after channel pruning over the original number of channels before channel pruning. From Fig.~\ref{fig:density}, the percentages of the remained channels automatically decided by our PCP has a similar trend with that in \cite{he2017channel}, which is carefully designed by human. Specifically, the percentages of the remained channels for deeper layers are higher. This observation is also reported in \cite{he2017channel}, which pointed out ``channel pruning gradually becomes hard, from shallower to deeper layers''. We also hypothesize that there are less redundant channels in deeper layers. 

\begin{table}[t]
\caption{Top-1 and Top-5 accuracies (\%) of different channel pruning methods when compressing AlexNet on the ImageNet dataset under different number of FLOPs and parameters. The Top-1 and Top-5 accuracies of AlexNet before channel pruning are 57.1\% and 79.9\%, respectively. We directly report the accuracies after the fine-tuning process.}
\begin{center}
\resizebox{0.48\textwidth}{!}{
\begin{tabular}{|c|c|c|c|c|}
\hline
Method & FLOPs (\%) & Parameters (\%) & Top-1 Acc. (\%) & Top-5 Acc. (\%) \\
\hline
\makecell{NISP-B \cite{yu2018nisp}} & 37.3 & 98.0 & - & 79.0 \\
\hline
\makecell{Channel Pruning~\cite{he2017channel} \\(our implementation)} & 38.8 & 96.3 & 55.7 & \makecell{78.9} \\
\hline
\textbf{Ours} & \makecell{\textbf{35.7}} & \textbf{96.8} & \textbf{56.9} & \makecell{\textbf{79.5}} \\
\hline
\end{tabular}}
\end{center}
\label{tab:aleximagenet}
\end{table}
\textbf{Results with AlexNet.}
We take AlexNet as another example to further demonstrate the effectiveness of our PCP method. AlexNet is an 8-layer network, which consists of 5 convolutional layers and 3 fully connected layers. Similar to the setting in VGG-16, we only prune the input of the last 4 convolutional layers (i.e., conv2 to conv5).

We empirically set the number of selected layers in the selecting step (i.e., $top\_n$) as 2. It takes 4 iterations to obtain the compressed model with 35.7\% FLOPs. In the fine-tuning stage, we set the initial learning rate as $2e^{-3}$ and use SGD optimizer with the batch size of 512 for optimization. Other settings are the same as those for pruning VGG-16 on ImageNet. We fine-tune the pruned AlexNet for 90 epochs, while it takes 90 epochs and 106 epochs in the baseline methods NISP-B~\cite{yu2018nisp} and Channel Pruning (CP)~\cite{he2017channel} to fine-tune the compressed model, respectively.

In Table \ref{tab:aleximagenet}, we compare the Top-1 and Top-5 accuracies of our PCP framework with different channel pruning methods. We compare our PCP method with the NISP-B method in \cite{yu2018nisp} because both methods prune the last 4 convolutional layers of AlexNet. NISP-B calculates the global channel importance scores and simultaneously prunes a set of channels from different layers with the smallest importance scores. When compared with the NISP-B method and the CP approach, it is clear that our PCP method achieves higher accuracy with a lower number of FLOPs.

\begin{table}[t]
\caption{Top-1 and Top-5 accuracies (\%) of different channel pruning methods when compressing ResNet-50 on the ImageNet dataset under different number of FLOPs and parameters. The Top-1 and Top-5 accuracies for ResNet-50 before channel pruning are 74.3\% and 92.1\%, respectively. We directly report the accuracies after the fine-tuning process.}
\begin{center}
\resizebox{0.48\textwidth}{!}{
\begin{tabular}{|c|c|c|c|c|}
\hline
Method & FLOPs (\%) & Parameters (\%) & Top-1 Acc. (\%) & Top-5 Acc. (\%) \\
\hline
\makecell{ThiNet-70 \cite{luoiccv2017}} & 63.2 & 66.3 & 72.0 & 90.7 \\
\hline
\makecell{ThiNet-50 \cite{luoiccv2017}} & 44.2 & 48.4 & 71.0 & 90.0 \\
\hline
\makecell{Channel Pruning \cite{he2017channel}} & 67.4 & 77.1 & 72.4 & 90.8 \\
\hline
\textbf{Ours} & \textbf{44.5} & \textbf{59.2} & \textbf{73.4} & \textbf{91.5} \\
\hline
\end{tabular}}
\end{center}
\label{tab:resnet50imagenet}
\end{table}

\begin{table*}[t]
\caption{Comparison of Top-1 Accuracies(\%) of different channel pruning methods when compressing the VGG-based DANN model on the Office-31 dataset under different compression ratios (CRs). We directly report the accuracies after the fine-tuning process.}
\small
\begin{center}
\setlength{\tabcolsep}{0.01\textwidth}{
\begin{tabular}{|c|c|c|c|c|c|c|c|}
\hline
CRs & Methods & A $\rightarrow$ W & W $\rightarrow$ A & A $\rightarrow$ D & D $\rightarrow$ A & D $\rightarrow$ W & W $\rightarrow$ D  \\
\hline
& \makecell{VGG16} & 70.0 & 56.4 & 74.1 & 53.5 & 95.7 & 99.0 \\
\hline
&\makecell{VGG16-DANN} & 81.9 & 64.3 & 80.9 & 63.4 & 97.1 & 99.6\\
\hline
2$\times$ & \makecell{CP \cite{he2017channel}} & 79.5 & 63.2 & 79.9 & 60.8 & 96.7 & 99.6 \\
& \makecell{PCP (w/o PL) (Ours)} & 80.2 & 63.5 & 80.8 & 62.8 & 96.8 & 99.6 \\
& \makecell{\textbf{PCP (w/ PL) (Ours)}} & \textbf{82.0} & \textbf{64.4} & \textbf{81.1} & \textbf{63.9} & \textbf{97.1} & \textbf{99.8} \\
\hline
4$\times$ & \makecell{CP \cite{he2017channel}} & 79.2 & 61.4 & 78.1 & 59.8 & 96.6 & 99.2 \\
& \makecell{PCP (w/o PL) (Ours)}  & 78.9 & 61.6 & 79.3 & 60.0 & 96.6 & 99.6 \\
& \makecell{\textbf{PCP (w/ PL) (Ours)}} & \textbf{81.8} & \textbf{64.4} & \textbf{80.9} & \textbf{63.6} & \textbf{97.1} & \textbf{99.6} \\
\hline
5$\times$ & \makecell{CP \cite{he2017channel}} & 76.2 & 60.7 & 77.9 & 58.4 & 95.9 & 98.8 \\
& \makecell{PCP (w/o PL) (Ours)} & 79.2 & 60.9 & 77.7 & 58.2 & 96.0 & 99.0 \\
& \makecell{\textbf{PCP (w/ PL) (Ours)}} & \textbf{81.6} & \textbf{64.3} & \textbf{80.7} & \textbf{63.6} & \textbf{97.0} & \textbf{99.6} \\
\hline
\end{tabular}
}
\end{center}
\label{table:uda_vgg}
\end{table*}

\begin{table*}[t]
\caption{Comparison of Top-1 Accuracies(\%) of different channel pruning methods when compressing the AlexNet-based DANN model on the Office-31 dataset under different compression ratios (CRs). We directly report the accuracies after the fine-tuning process. }
\small
\begin{center}
\setlength{\tabcolsep}{0.01\textwidth}{
\begin{tabular}{|c|c|c|c|c|c|c|c|}
\hline
CRs & Methods & A $\rightarrow$ W & W $\rightarrow$ A & A $\rightarrow$ D & D $\rightarrow$ A & D $\rightarrow$ W & W $\rightarrow$ D  \\
\hline
& \makecell{AlexNet} & 33.0 & 14.7 & 38.2 & 14.5 & 78.5 & 77.5 \\
\hline
&\makecell{AlexNet-DANN} & 	64.9&	47.9&	70.9&	50.8&	95.1&	97.8 \\
\hline
& \makecell{CP \cite{he2017channel}} &	61.9&	46.0&	66.9&	48.4&	93.8&	97.0 \\
2.7$\times$ & \makecell{PCP (w/o PL) (Ours)} & 62.4    &   47.0  & 66.9   & 48.6  & 95.0 & 97.2\\
& \makecell{\textbf{PCP (w/ PL) (Ours)}} & \textbf{64.0} & \textbf{48.1} & \textbf{68.9} & \textbf{51.3} & \textbf{95.1} & \textbf{97.6} \\
\hline
& \makecell{CP \cite{he2017channel}} &	60.7 & 43.8	& 65.1 & 45.8 &	92.7 & 96.5  \\
5$\times$ & \makecell{PCP (w/o PL) (Ours)} &  62.1 &  45.6  & 66.3 &  46.8 &  93.6 & 97.0 \\
& \makecell{\textbf{PCP (w/ PL) (Ours)}} & \textbf{63.5}  & \textbf{48.1}   & \textbf{69.1} & \textbf{51.2} &  \textbf{94.9} & \textbf{97.6}  \\
\hline
\end{tabular}
}
\end{center}
\label{table:uda_alex}
\end{table*}

\begin{table*}[t]
\caption{Comparison of Top-1 Accuracies(\%) of different channel pruning methods when compressing the ResNet50-based DANN model on the Office-31 dataset under different compression ratios (CRs). We directly report the accuracies after the fine-tuning process.}
\small
\begin{center}
\begin{tabular}{|c|c|c|c|c|c|c|c|}
\hline
CRs & Methods & A $\rightarrow$ W & W $\rightarrow$ A & A $\rightarrow$ D & D $\rightarrow$ A & D $\rightarrow$ W & W $\rightarrow$ D  \\
\hline
& \makecell{ResNet50} & 72.7 & 58.9 & 76.2 & 56.3 & 93.8 & 98.8 \\
\hline
&\makecell{ResNet50-DANN} & 82.6 & 64.8 & 81.1 & 63.9& 97.0& 99.4 \\
\hline
2$\times$ & \makecell{CP \cite{he2017channel}} &	80.6&	61.3&	79.7&	60.1&	95.9&	98.2 \\
& \makecell{PCP (w/o PL) (Ours)} & 80.8    &  62.4  & 80.3   & 61.3  & 96.6 & 98.8\\
& \makecell{\textbf{PCP (w/ PL) (Ours)}} & \textbf{83.0} & \textbf{64.5} & \textbf{82.3} & \textbf{64.4} &  \textbf{96.9}&\textbf{99.8}  \\
\hline
& \makecell{CP \cite{he2017channel}} &	71.8 & 56.9	& 74.9 & 52.2 & 94.0 & 96.8 \\
5$\times$ & \makecell{PCP (w/o PL) (Ours)} &  75.3  &   58.3 & 76.8 & 54.5  &  95.4 & 97.1 \\
& \makecell{\textbf{PCP (w/ PL) (Ours)}} & \textbf{78.4} & \textbf{61.2} & \textbf{79.1} & \textbf{56.9} & \textbf{96.1} & \textbf{98.4} \\
\hline
\end{tabular}
\end{center}
\label{table:uda_resnet}
\end{table*}

\textbf{Results with ResNet-50.}
We further perform the experiment to prune the ResNet-50 model~\cite{he2016deep} on the ImageNet dataset to demonstrate the effectiveness of our PCP framework. We empirically set the number of selected layers in the selecting step (i.e., $top\_n$) as 32 to accelerate our PCP framework and set the pre-defined number of channels to be pruned $\mathrm{f}^{(l)}_{t}$ as max\{30\% of the current number of channels, 40\}. It takes 2 iterations to obtain the compressed model with 44.5\% FLOPs. In the fine-tuning process, all settings are the same as those for pruning VGG-16 except that the batch size is set to 64. We fine-tune the pruned ResNet-50 for 10 epochs, which is the same as those in the baseline methods ThiNet~\cite{luoiccv2017} and Channel Pruning~\cite{he2017channel}.

Table \ref{tab:resnet50imagenet} shows the results of our PCP approach on the ImageNet dataset. We compare our method with two state-of-the-art methods: ThiNet~\cite{luoiccv2017} and Channel Pruning (CP)~\cite{he2017channel}. From Table \ref{tab:resnet50imagenet}, our PCP approach consistently outperforms the ThiNet method and the CP approach. Again, the above results clearly demonstrate the effectiveness of our PCP framework.

We also compare the running time of our PCP approach with the CP method for compressing the ResNet-50 model when using one GTX2080Ti GPU. For the CP method, it takes 1 hour for the pruning process and 19.3 hours for the fine-tuning process. In our PCP framework, it takes 3.4 hours for the pruning process (i.e., our three-step pipeline) and 19.3 hours for the fine-tuning process. The total running time of the whole framework increases by 11.8\% for our PCP framework when compared with that for the baseline CP method, which demonstrates that the extra time required by our PCP framework is limited.

\subsection{Experiments under unsupervised domain adaptation setting}
\textbf{Experiments on Office-31.}
Office-31 dataset \cite{saenko2010adapting} is a benchmark dataset to evaluate different domain adaptation methods for object recognition, which contains 4,110 real images from 31 classes. It contains three domains, including Amazon (A), Webcam (W) and DSLR (D). We follow the original DANN work \cite{ganin2015unsupervised} to use the common evaluation protocol for the domain adaptation task.

We perform the experiments by using AlexNet, VGG16 and ResNet50 as our backbone networks to extract features. The initial DANN model before channel pruning is trained by setting the learning rate as $5e^{-4}$ in the first iteration. We follow other deep transfer learning methods~\cite{ganin2015unsupervised, ganin2016domain, long2015learning} to gradually decrease the learning rate with the INV function. In this way, our learned initial DANN model achieves promising performance on the target domain, which provides a good starting point for model compression. To the best of our knowledge, the existing model compression methods are not specifically designed for compressing deep transfer learning methods like DANN. To this end, we design several baseline methods to demonstrate the effectiveness of our PCP method, which are listed below:

(1) VGG16 (\textit{resp.} AlexNet, ResNet50): We first pre-train the VGG16 (\textit{resp.} AlexNet, ResNet50) model by using the ImageNet dataset and then fine-tune the pre-trained VGG16 (\textit{resp.} AlexNet, ResNet50) model by only using labelled source samples in the Office-31 dataset. The results of the fine-tuned VGG16 (\textit{resp.} AlexNet, ResNet50) model are reported as VGG16 (\textit{resp.} AlexNet, ResNet50) in Table~\ref{table:uda_vgg} (\textit{resp.} Table~\ref{table:uda_alex}, Table~\ref{table:uda_resnet}).

(2) VGG16-DANN (\textit{resp.} AlexNet-DANN, ResNet50-DANN): We first pre-train the VGG16 (\textit{resp.} AlexNet, ResNet50) model by using the ImageNet dataset. Then we fine-tune the pre-trained VGG16 (\textit{resp.} AlexNet, ResNet50) model with the DANN method~\cite{ganin2016domain} by using all the labelled source samples and unlabelled target samples from the Office-31 dataset to obtain the initial DANN model to be pruned. The results are reported as VGG16-DANN (\textit{resp.} AlexNet-DANN, ResNet50-DANN) in Table~\ref{table:uda_vgg} (\textit{resp.} Table~\ref{table:uda_alex}, Table~\ref{table:uda_resnet}).

(3) CP~\cite{he2017channel}: As the source code from the channel pruning (CP) method~\cite{he2017channel} is publically available, we directly use this method to prune the channels of the initial DANN model by only using the labelled samples from the source domain in the Office-31 dataset. The results are reported as CP in Table~\ref{table:uda_vgg} (\textit{resp.} Table~\ref{table:uda_alex}, Table~\ref{table:uda_resnet}) when the backbone network is VGG16 (\textit{resp.} AlexNet, ResNet50).

(4) PCP (w/o PL) (Ours): We use our newly proposed PCP method to compress the initial DANN model by only using the labelled samples from the source domain in the Office-31 dataset and the results are reported as PCP (w/o PL) (Ours) in Table~\ref{table:uda_vgg} (\textit{resp.} Table~\ref{table:uda_alex}, Table~\ref{table:uda_resnet}) when the backbone network is VGG16 (\textit{resp.} AlexNet, ResNet50).

(5) PCP (w/ PL) (Ours): We use our PCP method to compress the initial DANN model by using both labelled source samples and pseudo-labelled target samples in the Office-31 dataset and the results are reported as PCP (w/ PL) (Ours) in Table~\ref{table:uda_vgg} (\textit{resp.} Table~\ref{table:uda_alex}, Table~\ref{table:uda_resnet}) when the backbone network is VGG16 (\textit{resp.} AlexNet, ResNet50).

\textbf{Results and Analysis.}
From the results in Table \ref{table:uda_vgg}, Table \ref{table:uda_alex} and Table \ref{table:uda_resnet}, we have the following observations: 
(1) VGG16-DANN, AlexNet-DANN and ResNet50-DANN models outperform the corresponding fine-tuned models VGG16, AlexNet and ResNet50, respectively, which demonstrates that it is beneficial to use the domain adaptation method DANN to explicitly reduce the data distribution mismatch between the source and the target domains before model compression.
(2) The performance of the compressed models after pruning the channels by using the CP method is often worse than the initial DANN model.
(3) Our method PCP (w/o PL) outperforms the CP method in most settings, which demonstrates that it is useful to use our newly proposed PCP method to prune the channels of the DANN model.
(4) Our method PCP (w/ PL) achieves the best performance under all of the six settings. The results for most tasks are as good as or even better than the initial DANN models, which indicates that it is beneficial to use pseudo-labelled target samples to compress models in our PCP approach under unsupervised domain adaptation setting.

\section{Conclusion}
In this work, we have proposed the Progressive Channel Pruning (PCP) framework to accelerate convolutional neural networks under both supervised learning and transfer learning settings. Our PCP method iteratively uses a three-step attempting-selecting-pruning pipeline to prune a small number of channels from several selected layers in each iteration. Our PCP approach can not only automatically decide the number of remained channels at each layer but also simultaneously obtain a series of compressed deep models at different compression ratios. It can also reduce the data distribution mismatch between different domains in the channel pruning process under unsupervised domain adaptation setting. Comprehensive experiments on two benchmark datasets demonstrate the effectiveness of our newly proposed PCP method under both settings.

As stated in Sec.~\ref{Sec:OurPCP}, the attempting step at each layer is independent from other layers in our PCP framework, which can be performed in parallel. In our future work, we will investigate the parallel implementation of the attempting step to further improve the efficiency of our PCP framework.

\ifCLASSOPTIONcaptionsoff
  \newpage
\fi



%


{\small
\bibliographystyle{IEEEtran}
\bibliography{egbib}
}

%
\newpage
\begin{IEEEbiography}[{\includegraphics[width=1in,height=1.25in,clip,keepaspectratio]{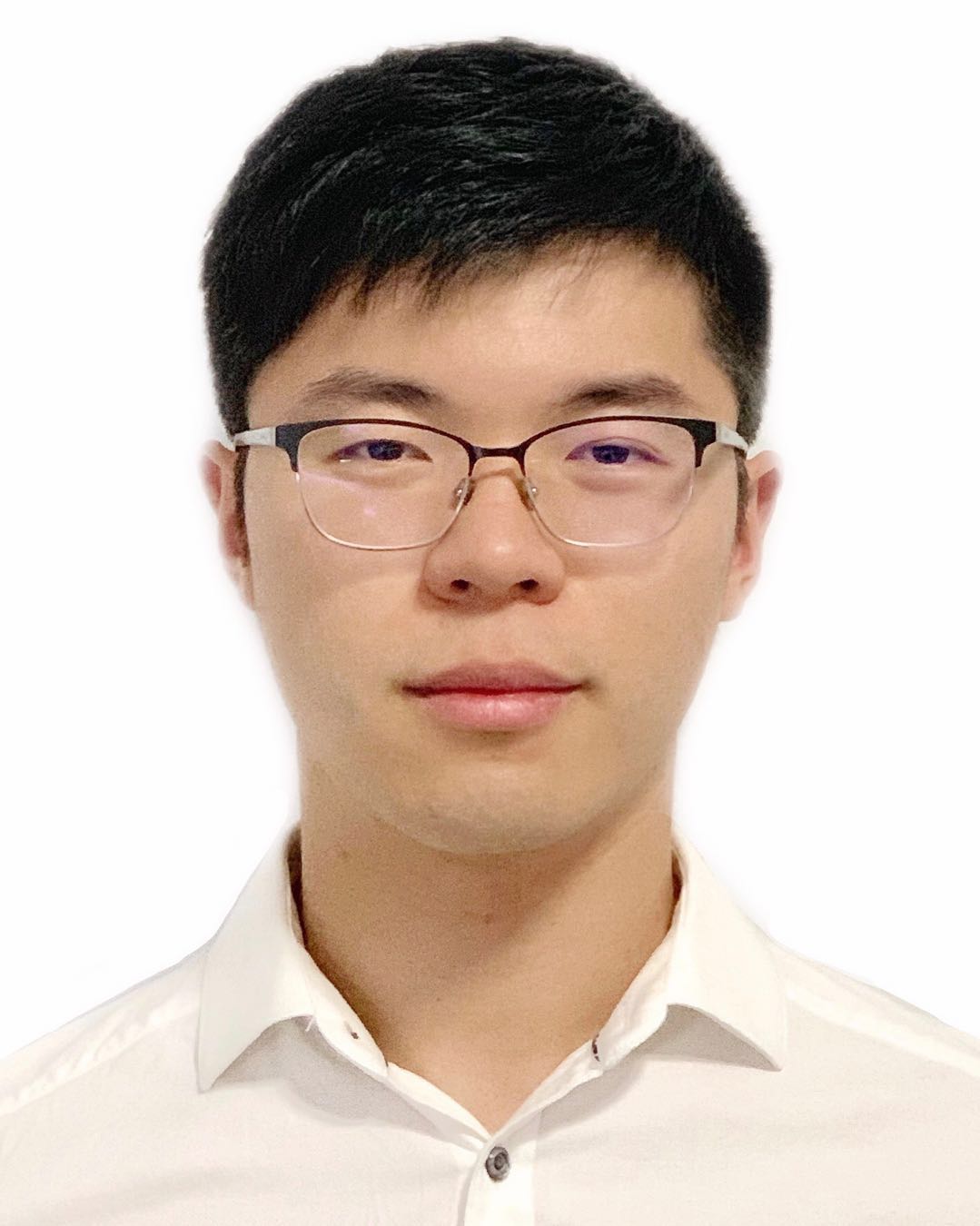}}]{Jinyang Guo}
received the BE degree in School of Electrical Engineering and Telecommunications from the University of New South Wales in 2017. He is currently pursuing the PhD degree in the School of Electrical and Information Engineering, the University of Sydney. His research interests include deep model compression and its applications on computer vision.
\end{IEEEbiography}
\vspace{-15mm}
\begin{IEEEbiography}
[{\includegraphics[width=1in,height=1.25in,clip,keepaspectratio]{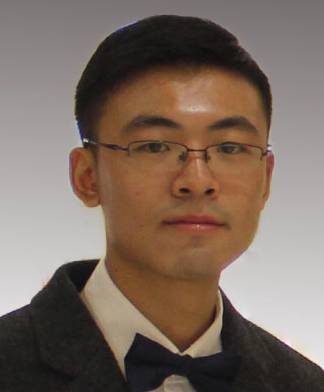}}]
{Weichen Zhang} received the BIT degree in School of Information Technology from the University of Sydney in 2016. He is currently working toward the PhD degree in the School of Electrical and Information Engineering, the University of Sydney. His current research interests include deep transfer learning and its applications in computer vision.
\end{IEEEbiography}
\vspace{-15mm}
\begin{IEEEbiography}[{\includegraphics[width=1in,height=1.25in,clip,keepaspectratio]{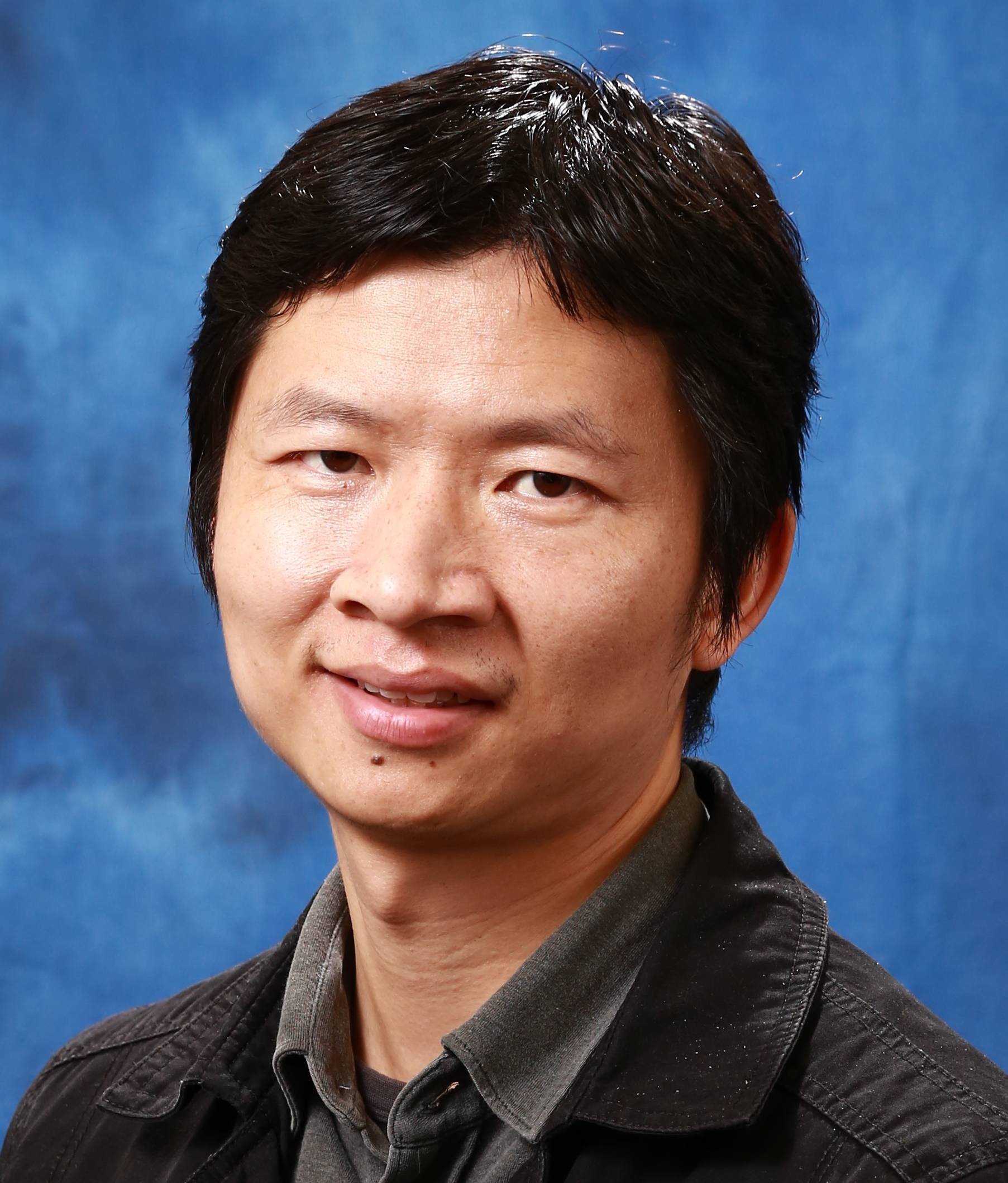}}]{Wanli Ouyang}
received the PhD degree in the Department of Electronic Engineering, Chinese University of Hong Kong. Since 2017, he is a senior lecturer with the University of Sydney. His research interests include image processing, computer vision, and pattern recognition. He is a senior member of the IEEE.
\end{IEEEbiography}
\vspace{-15mm}
\begin{IEEEbiography}[{\includegraphics[width=1in,height=1.25in,clip,keepaspectratio]{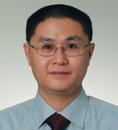}}]{Dong Xu}
 received the BE and PhD degrees from University of Science and Technology of China, in 2001 and 2005, respectively. While pursuing the PhD degree, he was an intern with Microsoft Research Asia, Beijing, China, and a research assistant with the Chinese University of Hong Kong, Shatin, Hong Kong, for more than two years. He was a post-doctoral research scientist with Columbia University, New York, NY, for one year. He worked as a faculty member with Nanyang Technological University, Singapore. Currently, he is a professor and chair in Computer Engineering with the School of Electrical and Information Engineering, the University of Sydney, Australia. His current research interests include computer vision, statistical learning, and multimedia content analysis. He was the co-author of a paper that won the Best Student Paper award in the IEEE Conferenceon Computer Vision and Pattern Recognition (CVPR) in 2010, and a paper that won the Prize Paper award in IEEE Transactions on Multimedia (T-MM) in 2014. He is a fellow of the IEEE.
\end{IEEEbiography}







\end{document}

%% file: definitions.tex
\usepackage{etoolbox}
\usepackage{bm}
\usepackage{amsthm}%
\usepackage{subfigure}%

\newcommand{\MyMapTemplatePrefixc}[4]{\expandafter#1\csname#3#4\endcsname{#2{#4}}} 
\forcsvlist{\MyMapTemplatePrefixc {\def} {\mathcal}{c}} {A,B,C,D,E,F,G,H,I,J,K,L,M,N,O,P,Q,R,S,T,U,V,W,X,Y,Z}  

\newcommand{\MyMapTemplatePrefixtb}[5]{\expandafter#1\csname#4#5\endcsname{#2{#3{#5}}}} 
\forcsvlist{\MyMapTemplatePrefixtb {\def} {\tilde}{\mathbf}{t}} {A,B,C,D,E,F,G,H,I,J,K,L,M,N,O,P,Q,R,S,T,U,V,W,X,Y,Z}  
\forcsvlist{\MyMapTemplatePrefixtb {\def} {\tilde}{\mathbf}{t}} {0,1,a,b,c,d,e,f,g,h,i,j,k,l,m,n,o,p,q,r,s,u,v,w,x,y,z}  

\newcommand{\MyMapTemplateNoPrefix}[3]{\expandafter#1\csname#3\endcsname{#2{#3}}}
\forcsvlist{\MyMapTemplateNoPrefix {\def} {\mathbf} } {0,1,a,b,c,d,e, f, g, h, i, j, k, l, m, n, o, p, q, r, u, v, w, x, y, z} 
\forcsvlist{\MyMapTemplateNoPrefix {\def} {\mathbf} } {A,B,C,D,E,F,G,H,I,J,K,L,M,N,O,P,Q,R,S,T,U,V,W,X,Y,Z}  

\usepackage{booktabs} 
\usepackage[table,dvipsnames]{xcolor}
\definecolor{rowblue}{RGB}{220,230,240}

%% file: main.bbl
\begin{thebibliography}{10}
\providecommand{\url}[1]{#1}
\csname url@samestyle\endcsname
\providecommand{\newblock}{\relax}
\providecommand{\bibinfo}[2]{#2}
\providecommand{\BIBentrySTDinterwordspacing}{\spaceskip=0pt\relax}
\providecommand{\BIBentryALTinterwordstretchfactor}{4}
\providecommand{\BIBentryALTinterwordspacing}{\spaceskip=\fontdimen2\font plus
\BIBentryALTinterwordstretchfactor\fontdimen3\font minus
  \fontdimen4\font\relax}
\providecommand{\BIBforeignlanguage}[2]{{%
\expandafter\ifx\csname l@#1\endcsname\relax
\typeout{** WARNING: IEEEtran.bst: No hyphenation pattern has been}%
\typeout{** loaded for the language `#1'. Using the pattern for}%
\typeout{** the default language instead.}%
\else
\language=\csname l@#1\endcsname
\fi
#2}}
\providecommand{\BIBdecl}{\relax}
\BIBdecl

\bibitem{he2017channel}
Y.~He, X.~Zhang, and J.~Sun, ``Channel pruning for accelerating very deep
  neural networks,'' in \emph{ICCV}, 2017, pp. 1398--1406.

\bibitem{luoiccv2017}
J.-H. Luo, J.~Wu, and W.~Lin, ``Thinet: A filter level pruning method for deep
  neural network compression,'' in \emph{ICCV}, 2017, pp. 5068--5076.

\bibitem{imagenet}
O.~Russakovsky, J.~Deng, H.~Su, J.~Krause, S.~Satheesh, S.~Ma, Z.~Huang,
  A.~Karpathy, A.~Khosla, M.~Bernstein \emph{et~al.}, ``Imagenet large scale
  visual recognition challenge,'' \emph{IJCV}, vol. 115, no.~3, pp. 211--252,
  2015.

\bibitem{saenko2010adapting}
K.~Saenko, B.~Kulis, M.~Fritz, and T.~Darrell, ``Adapting visual category
  models to new domains,'' in \emph{ECCV}, 2010, pp. 213--226.

\bibitem{ganin2016domain}
Y.~Ganin, E.~Ustinova, H.~Ajakan, P.~Germain, H.~Larochelle, F.~Laviolette,
  M.~Marchand, and V.~Lempitsky, ``Domain-adversarial training of neural
  networks,'' \emph{JMLR}, vol.~17, no.~59, pp. 1--35, 2016.

\bibitem{lebedev2014factorization}
V.~Lebedev, Y.~Ganin, M.~Rakhuba, I.~Oseledets, and V.~Lempitsky, ``Speeding-up
  convolutional neural networks using fine-tuned cp-decomposition,''
  \emph{arXiv preprint arXiv:1412.6553}, 2014.

\bibitem{jaderberg2014factorization}
M.~Jaderberg, A.~Vedaldi, and A.~Zisserman, ``Speeding up convolutional neural
  networks with low rank expansions,'' in \emph{BMVC}, 2014.

\bibitem{kim2015factorization}
Y.-D. Kim, E.~Park, S.~Yoo, T.~Choi, L.~Yang, and D.~Shin, ``Compression of
  deep convolutional neural networks for fast and low power mobile
  applications,'' \emph{arXiv preprint arXiv:1511.06530}, 2015.

\bibitem{gong2014factorization}
Y.~Gong, L.~Liu, M.~Yang, and L.~Bourdev, ``Compressing deep convolutional
  networks using vector quantization,'' \emph{arXiv preprint arXiv:1412.6115},
  2014.

\bibitem{xue2013factorization}
J.~Xue, J.~Li, and Y.~Gong, ``Restructuring of deep neural network acoustic
  models with singular value decomposition.'' in \emph{Interspeech}, 2013.

\bibitem{rastegari2016quantization}
M.~Rastegari, V.~Ordonez, J.~Redmon, and A.~Farhadi, ``Xnor-net: Imagenet
  classification using binary convolutional neural networks,'' in \emph{ECCV},
  2016, pp. 525--542.

\bibitem{guo2024compressing}
J.~Guo, J.~Wu, Z.~Wang, J.~Liu, G.~Yang, Y.~Ding, R.~Gong, H.~Qin, and X.~Liu,
  ``Compressing large language models by joint sparsification and
  quantization,'' in \emph{Forty-first International Conference on Machine
  Learning}, 2024.

\bibitem{lv2024ptq4sam}
C.~Lv, H.~Chen, J.~Guo, Y.~Ding, and X.~Liu, ``Ptq4sam: Post-training
  quantization for segment anything,'' in \emph{Proceedings of the IEEE/CVF
  Conference on Computer Vision and Pattern Recognition}, 2024, pp.
  15\,941--15\,951.

\bibitem{yang2024llmcbench}
G.~Yang, C.~He, J.~Guo, J.~Wu, Y.~Ding, A.~Liu, H.~Qin, P.~Ji, and X.~Liu,
  ``Llmcbench: Benchmarking large language model compression for efficient
  deployment,'' \emph{NeurIPS}, 2024.

\bibitem{bagherinezhad2017lcnn}
H.~Bagherinezhad, M.~Rastegari, and A.~Farhadi, ``Lcnn: Lookup-based
  convolutional neural network,'' in \emph{CVPR}, 2017.

\bibitem{lavin2016fast}
A.~Lavin and S.~Gray, ``Fast algorithms for convolutional neural networks,'' in
  \emph{CVPR}, 2016, pp. 4013--4021.

\bibitem{mathieu2014fast}
M.~Mathieu, M.~Henaff, and Y.~Lecun, ``Fast training of convolutional networks
  through ffts,'' in \emph{ICLR}, 2014.

\bibitem{vasilache2014fast}
N.~Vasilache, J.~Johnson, M.~Mathieu, S.~Chintala, S.~Piantino, and Y.~LeCun,
  ``Fast convolutional nets with fbfft: A gpu performance evaluation,''
  \emph{arXiv preprint arXiv:1412.7580}, 2014.

\bibitem{howard2017mobilenet}
A.~G. Howard, M.~Zhu, B.~Chen, D.~Kalenichenko, W.~Wang, T.~Weyand,
  M.~Andreetto, and H.~Adam, ``Mobilenets: Efficient convolutional neural
  networks for mobile vision applications,'' \emph{arXiv preprint
  arXiv:1704.04861}, 2017.

\bibitem{zhang2017shufflenet}
X.~Zhang, X.~Zhou, M.~Lin, and J.~Sun, ``Shufflenet: An extremely efficient
  convolutional neural network for mobile devices.'' in \emph{CVPR}, 2018.

\bibitem{liu2024lta-pcs}
J.~Liu, J.~Li, K.~Wang, H.~Guo, J.~Yang, J.~Peng, K.~Xu, X.~Liu, and J.~Guo,
  ``Lta-pcs: Learnable task-agnostic point cloud sampling,'' in \emph{CVPR},
  2024.

\bibitem{han2016eie}
S.~Han, X.~Liu, H.~Mao, J.~Pu, A.~Pedram, M.~A. Horowitz, and W.~J. Dally,
  ``{EIE}: Efficient inference engine on compressed deep neural network,'' in
  \emph{Computer Architecture (ISCA), 2016 ACM/IEEE 43rd Annual International
  Symposium on}, 2016, pp. 243--254.

\bibitem{hu2016nettrim}
H.~Hu, R.~Peng, Y.-W. Tai, and C.-K. Tang, ``Network trimming: A data-driven
  neuron pruning approach towards efficient deep architectures,'' \emph{arXiv
  preprint arXiv:1607.03250}, 2016.

\bibitem{li2017filter}
H.~Li, A.~Kadav, I.~Durdanovic, H.~Samet, and H.~P. Graf, ``Pruning filters for
  efficient convnets,'' \emph{ICLR}, 2016.

\bibitem{molchanov2017taylor}
P.~Molchanov, S.~Tyree, T.~Karras, T.~Aila, and J.~Kautz, ``Pruning
  convolutional neural networks for resource efficient inference,''
  \emph{ICLR}, 2017.

\bibitem{han2015learning}
S.~Han, J.~Pool, J.~Tran, and W.~Dally, ``Learning both weights and connections
  for efficient neural network,'' in \emph{NIPS}, 2015, pp. 1135--1143.

\bibitem{guo2020multi}
J.~Guo, W.~Ouyang, and D.~Xu, ``Multi-dimensional pruning: A unified framework
  for model compression,'' in \emph{CVPR}, 2020.

\bibitem{guo2023multidimensional}
J.~Guo, D.~Xu, and W.~Ouyang, ``Multidimensional pruning and its extension: A
  unified framework for model compression,'' \emph{IEEE Transactions on Neural
  Networks and Learning Systems}, 2023.

\bibitem{guo2020channel}
J.~Guo, W.~Ouyang, and D.~Xu, ``Channel pruning guided by classification loss
  and feature importance,'' in \emph{AAAI}, 2020.

\bibitem{wang2024ptsbench}
Z.~Wang, J.~Guo, R.~Gong, Y.~Yong, A.~Liu, Y.~Huang, J.~Liu, and X.~Liu,
  ``Ptsbench: A comprehensive post-training sparsity benchmark towards
  algorithms and models,'' in \emph{ACM Multimedia 2024}.

\bibitem{guo2021jointpruning}
J.~Guo, J.~Liu, and D.~Xu, ``Jointpruning: Pruning networks along multiple
  dimensions for efficient point cloud processing,'' \emph{IEEE Transactions on
  Circuits and Systems for Video Technology}, 2021.

\bibitem{guo2023cbanet}
J.~Guo, D.~Xu, and G.~Lu, ``Cbanet: Towards complexity and bitrate adaptive
  deep image compression using a single network,'' \emph{IEEE Transactions on
  Image Processing}, 2023.

\bibitem{guo20223d}
J.~Guo, J.~Liu, and D.~Xu, ``3d-pruning: A model compression framework for
  efficient 3d action recognition,'' \emph{IEEE Transactions on Circuits and
  Systems for Video Technology}, vol.~32, no.~12, pp. 8717--8729, 2022.

\bibitem{dentonnips2014}
E.~L. Denton, W.~Zaremba, J.~Bruna, Y.~LeCun, and R.~Fergus, ``Exploiting
  linear structure within convolutional networks for efficient evaluation,'' in
  \emph{NIPS}, 2014, pp. 1269--1277.

\bibitem{girshickcvpr2015fastrcnn}
R.~Girshick, ``Fast {R-CNN},'' in \emph{ICCV}, 2015, pp. 1440--1448.

\bibitem{zhao2019variational}
C.~Zhao, B.~Ni, J.~Zhang, Q.~Zhao, W.~Zhang, and Q.~Tian, ``Variational
  convolutional neural network pruning,'' in \emph{CVPR}, 2019.

\bibitem{lin2019towards}
S.~Lin, R.~Ji, C.~Yan, B.~Zhang, L.~Cao, Q.~Ye, F.~Huang, and D.~Doermann,
  ``Towards optimal structured cnn pruning via generative adversarial
  learning,'' in \emph{CVPR}, 2019.

\bibitem{peng2019collaborative}
H.~Peng, J.~Wu, S.~Chen, and J.~Huang, ``Collaborative channel pruning for deep
  networks,'' in \emph{ICML}, 2019.

\bibitem{heeccv2018}
Y.~He, J.~Lin, Z.~Liu, H.~Wang, L.-J. Li, and S.~Han, ``{AMC}: Automl for model
  compression and acceleration on mobile devices,'' in \emph{ECCV}, 2018, pp.
  815--832.

\bibitem{chen2019cooperative}
S.~Chen, W.~Wang, and S.~J. Pan, ``Cooperative pruning in cross-domain deep
  neural network compression,'' in \emph{IJCAI}.\hskip 1em plus 0.5em minus
  0.4em\relax AAAI Press, 2019.

\bibitem{long2015learning}
M.~Long, Y.~Cao, J.~Wang, and M.~Jordan, ``Learning transferable features with
  deep adaptation networks,'' in \emph{ICML}, 2015, pp. 97--105.

\bibitem{long2016unsupervised}
M.~Long, H.~Zhu, J.~Wang, and M.~I. Jordan, ``Unsupervised domain adaptation
  with residual transfer networks,'' in \emph{NIPS}, 2016, pp. 136--144.

\bibitem{sun2016deep}
B.~Sun and K.~Saenko, ``Deep coral: Correlation alignment for deep domain
  adaptation,'' in \emph{ECCV}, 2016, pp. 443--450.

\bibitem{tzeng2014deep}
E.~Tzeng, J.~Hoffman, N.~Zhang, K.~Saenko, and T.~Darrell, ``Deep domain
  confusion: Maximizing for domain invariance,'' \emph{arXiv preprint
  arXiv:1412.3474}, 2014.

\bibitem{bousmalis2017unsupervised}
K.~Bousmalis, N.~Silberman, D.~Dohan, D.~Erhan, and D.~Krishnan, ``Unsupervised
  pixel-level domain adaptation with generative adversarial networks,'' in
  \emph{CVPR}, 2017, pp. 3722--3731.

\bibitem{bousmalis2016domain}
K.~Bousmalis, G.~Trigeorgis, N.~Silberman, D.~Krishnan, and D.~Erhan, ``Domain
  separation networks,'' in \emph{NIPS}, 2016, pp. 343--351.

\bibitem{ganin2015unsupervised}
Y.~Ganin and V.~Lempitsky, ``Unsupervised domain adaptation by
  backpropagation,'' in \emph{ICML}, 2015, pp. 1180--1189.

\bibitem{liu2016coupled}
M.-Y. Liu and O.~Tuzel, ``Coupled generative adversarial networks,'' in
  \emph{NIPS}, 2016, pp. 469--477.

\bibitem{tzeng2017adversarial}
E.~Tzeng, J.~Hoffman, K.~Saenko, and T.~Darrell, ``Adversarial discriminative
  domain adaptation,'' in \emph{CVPR}, 2017, pp. 2962--2971.

\bibitem{zhang2018collaborative}
W.~Zhang, W.~Ouyang, W.~Li, and D.~Xu, ``Collaborative and adversarial network
  for unsupervised domain adaptation,'' in \emph{CVPR}, 2018, pp. 3801--3809.

\bibitem{simonyan2014very}
K.~Simonyan and A.~Zisserman, ``Very deep convolutional networks for
  large-scale image recognition,'' \emph{arXiv}, 2014.

\bibitem{he2016deep}
K.~He, X.~Zhang, S.~Ren, and J.~Sun, ``Deep residual learning for image
  recognition,'' in \emph{CVPR}, 2016, pp. 770--778.

\bibitem{krizhevsky2012imagenet}
A.~Krizhevsky, I.~Sutskever, and G.~E. Hinton, ``Imagenet classification with
  deep convolutional neural networks,'' in \emph{NIPS}, 2012, pp. 1097--1105.

\bibitem{yu2018nisp}
R.~Yu, A.~Li, C.-F. Chen, J.-H. Lai, V.~I. Morariu, X.~Han, M.~Gao, C.-Y. Lin,
  and L.~S. Davis, ``{NISP}: Pruning networks using neuron importance score
  propagation,'' in \emph{CVPR}, 2018.

\end{thebibliography}
